\documentclass{article}

\PassOptionsToPackage{numbers, compress}{natbib}
\usepackage[final]{neurips_2025}

\usepackage[utf8]{inputenc} % allow utf-8 input
\usepackage[T1]{fontenc}    % use 8-bit T1 fonts
\usepackage{hyperref}       % hyperlinks
\usepackage{url}            % simple URL typesetting
\usepackage{booktabs}       % professional-quality tables
\usepackage{amsfonts}       % blackboard math symbols
\usepackage{nicefrac}       % compact symbols for 1/2, etc.
\usepackage{microtype}      % microtypography
\usepackage{xcolor}         % colors
\usepackage{graphicx}
\usepackage{subcaption}
\usepackage{booktabs}
\usepackage{multirow}
\usepackage{colortbl}
\usepackage[capitalize,noabbrev,nameinlink]{cleveref}

\definecolor{mydarkblue}{rgb}{0,0.08,0.45}
\hypersetup{
    colorlinks=true,
    citecolor=mydarkblue,
    urlcolor=mydarkblue,
    linkcolor = mydarkblue
}

\title{Longer Context, Deeper Thinking: Uncovering the Role of Long-Context Ability in Reasoning}

\author{
    \textbf{Wang Yang$^1$, Zirui Liu$^2$, Hongye Jin$^3$, Qingyu Yin}\\ 
    \textbf{Vipin Chaudhary$^1$, Xiaotian Han$^1$}\\
    $^1$Case Western Reserve University $^2$ University of Minnesota - Twin Cities\\ $^3$Texas A\&M University \\
    \texttt{\{wxy320,vipin,xhan\}@case.edu, zrliu@umn.edu, jhy0410@tamu.edu} \\
}

\begin{document}

\maketitle

\begin{abstract}
    Recent language models exhibit strong reasoning capabilities, yet the influence of long-context capacity on reasoning remains underexplored. In this work, we hypothesize that current limitations in reasoning stem, in part, from insufficient long-context capacity, motivated by empirical observations such as \textit{i)} higher context window length often leads to stronger reasoning performance, and \textit{ii)} failed reasoning cases resemble failed long-context cases. To test this hypothesis, we examine whether enhancing a model’s long-context ability before Supervised Fine-Tuning (SFT) leads to improved reasoning performance. Specifically, we compared models with identical architectures and fine-tuning data but varying levels of long-context capacity. Our results reveal a consistent trend: models with stronger long-context capacity achieve significantly higher accuracy on reasoning benchmarks after SFT. Notably, these gains persist even on tasks with short input lengths, indicating that long-context training offers generalizable benefits for reasoning performance. These findings suggest that long-context modeling is not just essential for processing lengthy inputs, but also serves as a critical foundation for reasoning. We advocate for treating long-context capacity as a first-class objective in the design of future language models. Our code is anonymously available at \url{https://github.com/uservan/LCTMerge}.
\end{abstract}

\begin{figure}[h]
    \centering
    \begin{subfigure}{0.199\textwidth}
        \centering\includegraphics[height=1.5in]{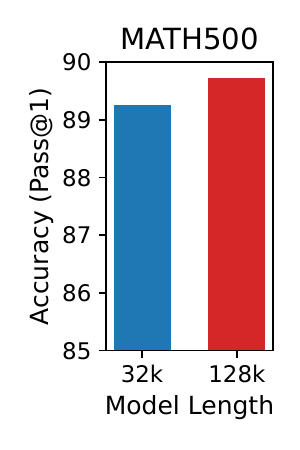}
    \end{subfigure}
    \begin{subfigure}{0.199\textwidth}
        \centering
        \includegraphics[height=1.5in]{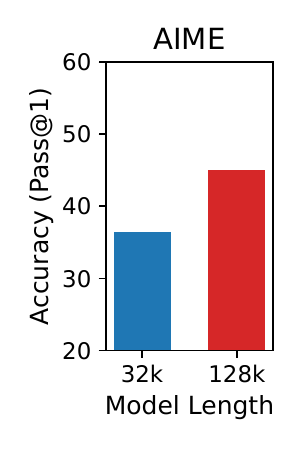}
    \end{subfigure}
    \begin{subfigure}{0.55\textwidth}
        \centering
        \includegraphics[height=1.45in]{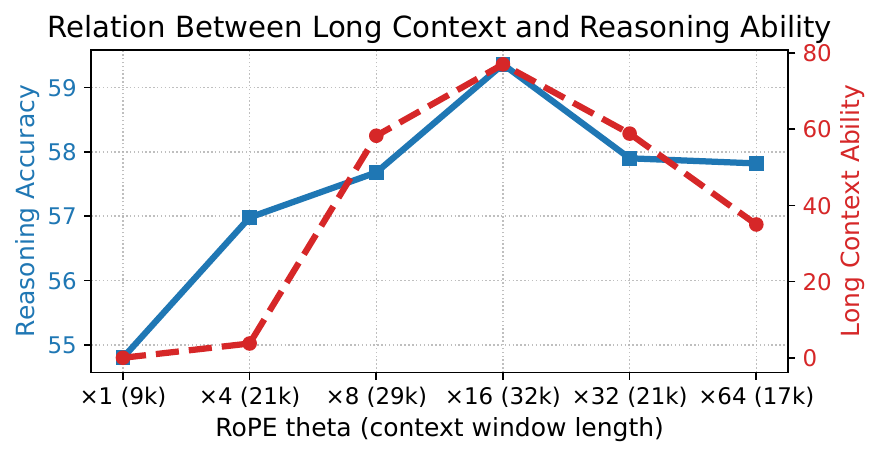}
    \end{subfigure}\vspace{-5pt}
    \caption{Impact of long-context capacity on mathematical reasoning. \textbf{Left:} Accuracy (Pass@1) on MATH500 and AIME datasets for public models with 32k and 128k context lengths, showing consistent improvements in reasoning performance with longer context windows. The 32k and 128k LLMs refer to three different public models, as shown in \cref{tab:Performance comparison}. \textbf{Right:} Reasoning accuracy versus RoPE theta values, highlighting a strong correlation between long-context capacity and reasoning performance. Increasing the RoPE theta value typically extends the effective context window length.}
    \label{fig:intro}
\end{figure}

\section{Introduction}
Large language models (LLMs) have recently demonstrated impressive reasoning capabilities across a wide range of benchmarks~\citep{guo2025deepseek, team2025kimi, abdin2025phi4reasoningtechnicalreport}. Despite this progress, the underlying factors that influence such reasoning abilities remain only partially understood. One particularly underexplored dimension is long-context ability—the model's capacity to utilize a longer reasoning path during inference—which could affect the reasoning performances and the reasoning model fine-tuning.

While prior work has primarily focused on training paradigms and dataset quality to enhance reasoning~\citep{deepmath,openthoughts,moshkov2025aimo2}, we hypothesize that a model’s reasoning ability is also fundamentally constrained by its long-context capacity. This hypothesis is grounded in three empirical observations. \textit{First}, models with extended context lengths (e.g., 128k vs. 32k) consistently achieve higher accuracy on reasoning benchmarks such as MATH500 and AIME, suggesting a direct performance benefit from stronger long-context modeling (\cref{tab:Performance comparison}, \cref{fig:intro}). \textit{Second}, case studies reveal that failed generations often involve extremely long outputs, sometimes truncated at generation limits, and exhibit issues like repetition or incorrect cross-referencing—failure patterns strongly linked to inadequate long-context capability (\cref{fig:case study}, \cref{fig:case study2}). \textit{Third}, modern reasoning datasets now include many samples exceeding 8K or even 10K tokens—substantially longer than early CoT data—requiring models to learn from long, variable-length reasoning sequences (\cref{fig: token distribution and length}). Together, these findings underscore long-context capacity as a critical factor for reasoning ability, and motivate the central question: \textit{Does improving a model’s long-context ability during pretraining enhance downstream reasoning?}

To rigorously investigate this, we conduct a controlled study comparing language models with identical architectures and fine-tuning data, but varying degrees of long-context pretraining. Our experimental results reveal a consistent and compelling trend: models with stronger long-context capabilities consistently outperform their counterparts on reasoning tasks after SFT. Notably, these improvements extend to reasoning problems with short input lengths, suggesting that long-context training imparts generalizable cognitive benefits that go beyond simply processing long sequences. As shown in \cref{fig:intro}, \texttt{LLaMA3-8B-Instruct} exhibits varying reasoning performance after training when equipped with different levels of long-context capability. Notably, reasoning ability tends to increase or decrease in accordance with the strength of the model’s long-context capacity, suggesting a direct correlation between the model's reasoning ability and long-context capacity.  

Based on our experimental results, we thus propose a \textit{Recipe for Reasoning Fine-Tuning}, which advocates for \textit{appropriately enhancing a model’s long-context capacity prior to reasoning SFT}—for instance, by extending its context length to 128K tokens. Applying this recipe to \texttt{Qwen2.5-Math-7B-Instruct}, we observe substantial improvements: performance on \texttt{MATH500} increases from an average of $85.04$ to $88.70$, and on \texttt{AIME}, from $15.00$ to $28.00$.

\section{Motivation: behavioral evidence suggests a connection between long context and reasoning}
In this section, we present a set of empirical observations that suggest a strong behavioral connection between a model’s long-context capability and its reasoning performance. Through controlled comparisons, token length analyses, and failure case studies, we observe that LLMs with stronger long-context abilities not only perform better on reasoning benchmarks but also handle diverse and extended reasoning sequences more reliably. These findings collectively highlight long-context modeling as a key factor and component in enabling strong reasoning ability.

\subsection{Higher context window length often leads to stronger reasoning performance.}

Recent advances in long-context modeling have enabled language models (LLMs) to process substantially longer sequences. However, it remains unclear whether such long-context capacities yield tangible benefits for reasoning tasks. In this section, we empirically examine the relationship between long-context ability and reasoning performance. We collect a set of well-known open-source reasoning models fine-tuned from \texttt{Qwen/Qwen2.5-7B-Instruct}. These models are categorized into two groups based on their long-context capacity: 32k and 128k tokens. We then evaluate and compare their reasoning performance on two math reasoning benchmarks: \texttt{MATH500} and \texttt{AIME}. 

The detailed results are reported in \cref{tab:Performance comparison}. \cref{fig:intro} presents that models with longer context lengths (128k vs. 32k) consistently achieve higher accuracy on mathematical reasoning benchmarks such as MATH500 and AIME. This suggests that the ability to encode and maintain longer contextual dependencies can directly translate into better reasoning capabilities. These results collectively highlight the importance of effective long-context training—not only for tasks involving long inputs, but also for general reasoning even when test-time inputs are relatively short.

\begin{table}[ht]
\caption{Performance comparison on MATH500 and AIME benchmarks for some popular open-source reasoning models with different long-context abilities at 32k and 128k context lengths. Reasoning models with enhanced long-context capacity (128k) generally exhibit improved performance, particularly on the AIME benchmark. Averages are reported in the bottom row.}
\label{tab:Performance comparison}
\resizebox{\textwidth}{!}{
\begin{tabular}{@{}c|cc|c|cc@{}}
\toprule
Long Context Ability at 32k   & MATH500 & AIME   & Long Context Ability at 128k             & MATH500 & AIME   \\ \midrule
OpenR1-Qwen-7B     & 90.36   & 43.11  & DeepSeek-R1-Distill-Qwen-7B & 91.68   & 45.56  \\
OpenThinker-7B     & 86.80   & 25.78  & OpenMath-Nemotron-7B        & 94.00   & 74.67  \\
OpenThinker2-7B    & 90.60   & 40.22  & DeepMath-Zero-7B            & 83.48   & 14.67  \\
OpenThinker3-7B    & 93.72   & 64.00  & AceReason-Nemotron-7B       & 93.84   & 62.89  \\ \midrule
Avg (32k)          & 90.37   & 43.28  & Avg (128k)                  & 90.75   & 49.45  \\ \bottomrule
\end{tabular}
}
\end{table}
\begin{figure}[ht]
    \centering
    
    \begin{subfigure}{1.0\textwidth}
        \centering
        \includegraphics[width=\linewidth]{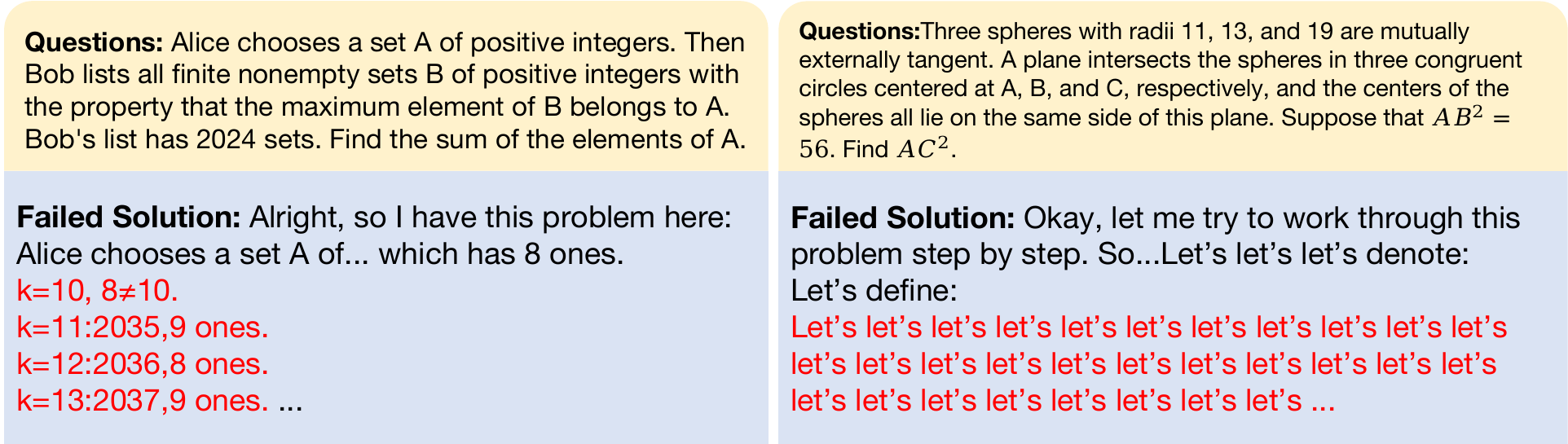}
    \end{subfigure}
    \vspace{-14pt}
    \caption{
    \textbf{Case Study: Repetition Failure.} Two failure cases where the model produces clearly repetitive sentences in its answers. Such repetition is a common symptom of insufficient long-context capability, leading to strange responses and degraded reasoning quality in extended sequences.
    }
    \label{fig:case study}
\end{figure}
\begin{figure}[ht]
    \centering
    
    \begin{subfigure}{1.0\textwidth}
        \centering
        \includegraphics[width=\linewidth]{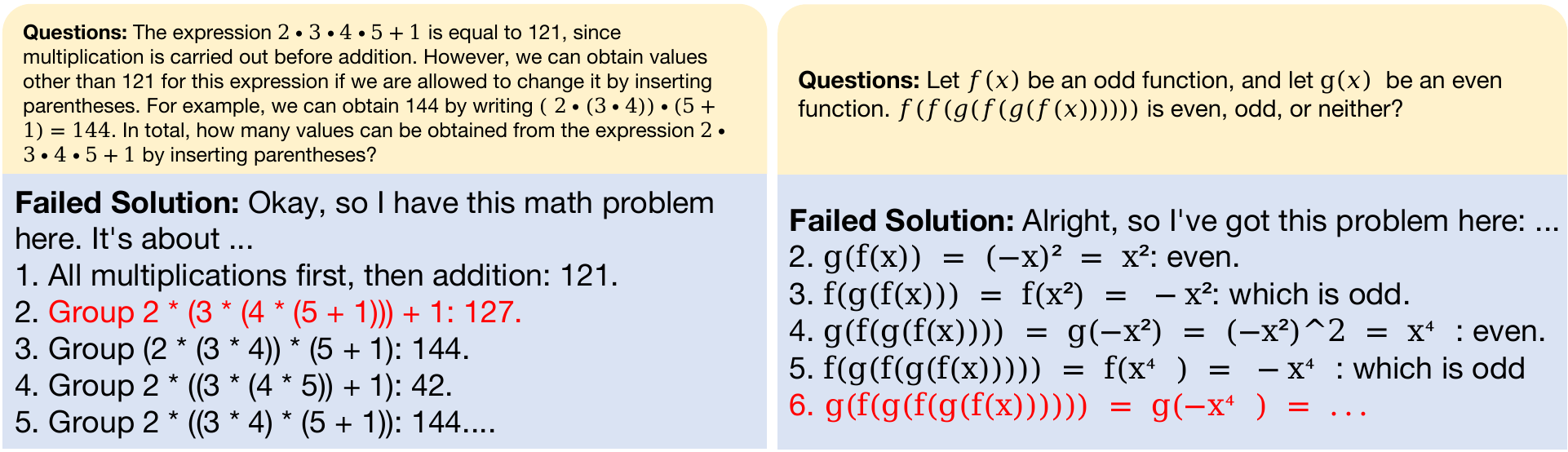}
    \end{subfigure}
    \vspace{-14pt}
    \caption{
    \textbf{Case Study: Contextual Reference Failures.} Two failure cases where the model makes incorrect references to expressions introduced earlier in the problem. These errors occur in the later stages of the response and reflect a typical symptom of insufficient long-context capability.
    }
    \label{fig:case study2}
\end{figure}

\subsection{Case study: failed reasoning cases resemble failed long-context cases.}
\begin{figure}[t]
    \centering
    
    \begin{subfigure}{0.32\textwidth}
        \centering
        \includegraphics[width=\linewidth]{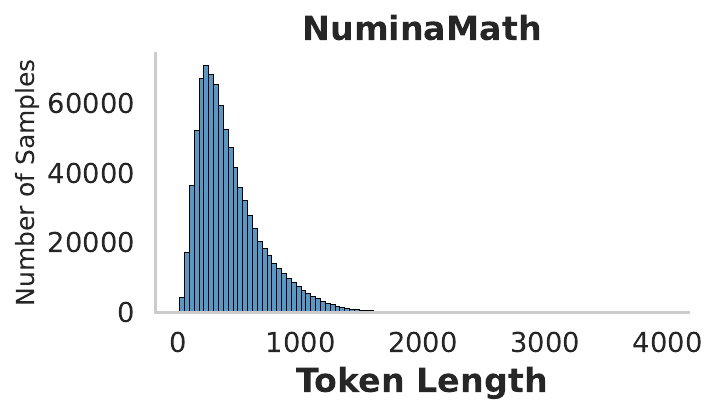}
    \end{subfigure}
    \begin{subfigure}{0.32\textwidth}
        \centering
        \includegraphics[width=\linewidth]{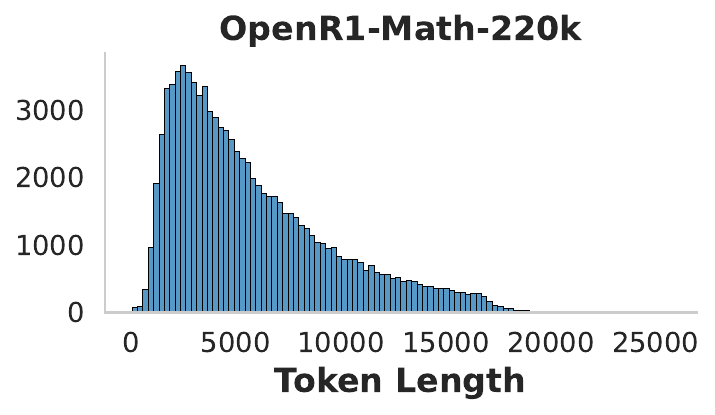}
        % \caption{OpenR1-Math-220k}
    \end{subfigure}
    \begin{subfigure}{0.32\textwidth}
        \centering
        \includegraphics[width=\linewidth]{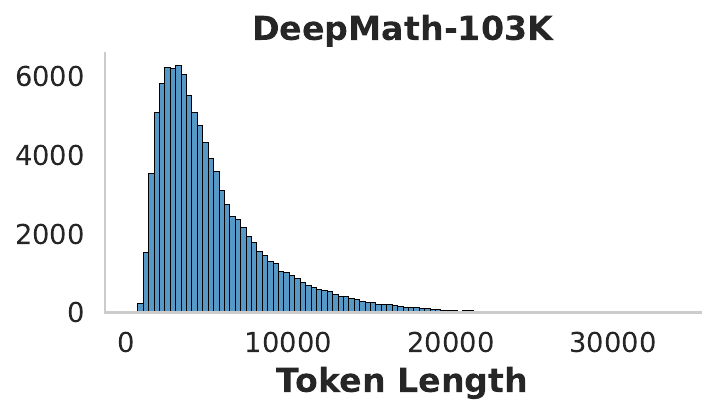}
    \end{subfigure}
        \begin{subfigure}{0.32\textwidth}
        \centering
        \includegraphics[width=\linewidth]{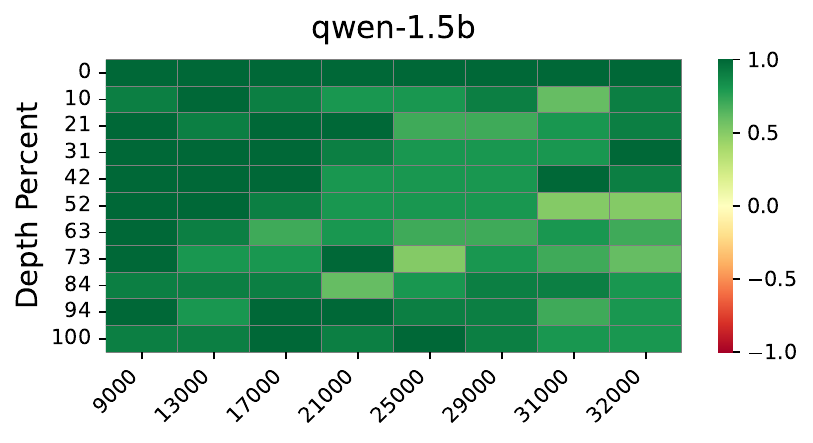}
    \end{subfigure}
    \begin{subfigure}{0.32\textwidth}
        \centering
        \includegraphics[width=\linewidth]{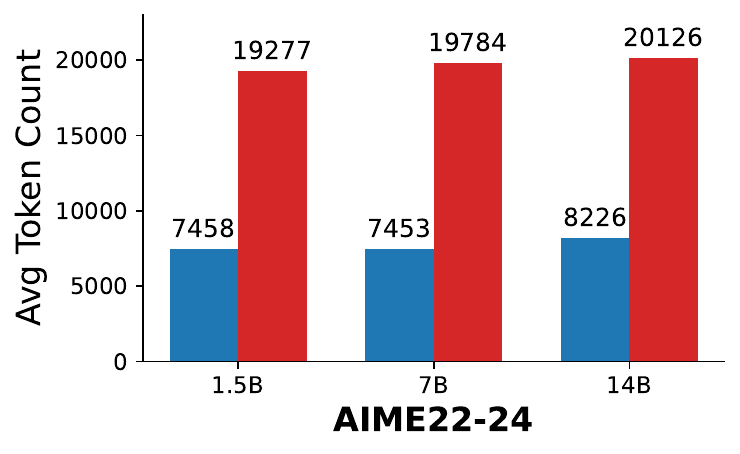}
    \end{subfigure}
    \begin{subfigure}{0.32\textwidth}
        \centering
        \includegraphics[width=\linewidth]{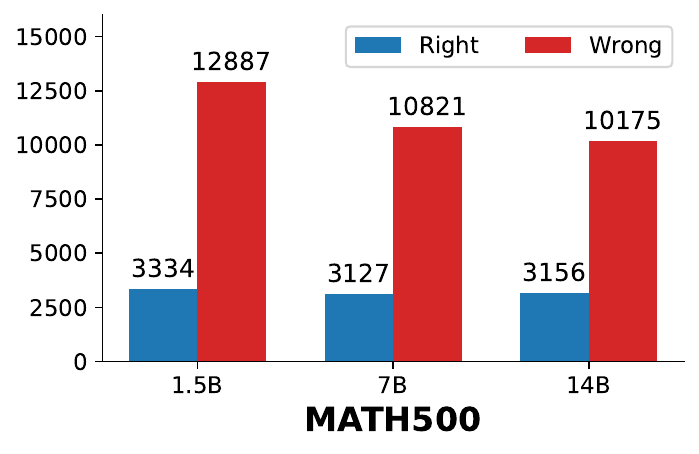}
    \end{subfigure}
    \vspace{-5pt}
    \caption{
    \textbf{Top:} Length distribution of three reasoning datasets. \texttt{NuminaMath-CoT} represents early chain-of-thought (CoT) data with short sequences, while \texttt{OpenR1-Math-220K} and \texttt{DeepMath-103K}, generated by \texttt{DeepSeek-R1}, exhibit significantly longer outputs.
    \textbf{Bottom-left:} Performance of \texttt{DeepSeek-Distilled-Qwen-1.5B} on the \textit{Needle-in-a-Haystack} benchmark with 32K context.
    \textbf{Bottom-middle/right:} Average output lengths of correct and incorrect generations on \texttt{AIME} and \texttt{MATH500} for \texttt{DeepSeek-Distilled-Qwen-1.5B}, \texttt{7B}, and \texttt{14B}. Incorrect answers consistently exhibit longer output lengths, indicating potential limitations of long-context ability in reasoning.
    }
    \label{fig: token distribution and length}
\end{figure}

We analyze failed cases of reasoning models and long-context inputs, revealing a strong connection between reasoning ability and long-context capacity. Specifically, we examine the average input lengths of correct and incorrect predictions made by three reasoning models—\texttt{DeepSeek-R1-Distill-Qwen-1.5B}, \texttt{7B}, and \texttt{14B}—on two math benchmarks: \texttt{MATH500} and \texttt{AIME}. As shown in \cref{fig: token distribution and length}, incorrect predictions are typically associated with inputs exceeding 10k tokens, suggesting that these failures may stem from insufficient long-context handling. 

To further investigate this hypothesis, we manually inspect a subset of long-output failures and identify two recurring patterns, repetition and contextual reference failures, 
due to limited long-context capability. The first pattern involves excessive repetition, where the model loops over the same sentence or phrase, failing to advance the solution, as shown in  \cref{fig:case study}. The second pattern arises in the latter part of the output, where the model incorrectly recalls earlier mathematical expressions from the problem, leading to flawed reasoning and incorrect conclusions, as shown in \cref{fig:case study2}. Both failure modes underscore the model’s struggle to maintain coherence and accuracy over long sequences—a well-known limitation of inadequate long-context capacity that degrades and distorts in-context learning performance in LLMs.

\subsection{Long and variable reasoning data necessitate long-context models.}

Current reasoning models are typically fine-tuned on long chain-of-thought (CoT) datasets generated by large-scale reasoning models. A key characteristic of these models is their tendency to produce \textit{long and variable-length outputs}. As a result, the resulting datasets exhibit broad length distributions, with many examples exceeding 10K tokens—far longer than early CoT-style data. This necessitates that fine-tuned models be capable of handling such long and diverse sequences. In \cref{fig: token distribution and length}, we analyze the length distributions of three representative datasets: \texttt{NuminaMath-CoT}, \texttt{OpenR1-Math-220K}, and \texttt{DeepMath-103K}. \texttt{NuminaMath-CoT}, representing early CoT-style data, primarily contains samples under 1K tokens. In contrast, \texttt{OpenR1-Math-220K} and \texttt{DeepMath-103K}, both collected from \texttt{DeepSeek-R1}, contain significantly longer reasoning sequences, with a large proportion of samples exceeding 4K tokens and some even surpassing 10K tokens, which is totally different with early CoT datasets like \texttt{NuminaMath-CoT}. 

There is a lack of research exploring how such increased sequence lengths interact with or depend on the model's long-context capability after the release of Deepseek-R1. For instance, if a model lacks sufficient long-context ability, training on long reasoning sequences may fail to yield expected improvements—or even negatively impact performance—due to the model’s inability to effectively utilize the full input. While modern models are advertised to handle long sequences (e.g., \texttt{Qwen2.5-1.5B-Instruct} supports up to 32K tokens), their effective context length is often substantially shorter. As shown in \cref{fig: token distribution and length}, we evaluate \texttt{Qwen2.5-1.5B-Instruct} on the \textit{Needle-in-a-Haystack} benchmark and observe that the model fails to maintain high accuracy across all cases in 32k contexts, indicating its limitations in effective long-context processing and long-context ability.

\begin{table}[t]
\caption{
Effective context length and long-context benchmark performance of \texttt{LLaMA3-8B-Instruct} under different \texttt{RoPE theta} scaling factors. 
We report the estimated effective context length, Needle-in-a-Haystack (NIAH) retrieval accuracy at 32k, as well as performance on LongBench and RULER. 
Results show that scaling up to RoPE $\times$16 consistently improves long-context robustness across all benchmarks, but further scaling (e.g., $\times$32 and $\times$64) leads to diminishing or even negative returns. 
Notably, the effective length surpasses the maximum sequence length (16k) of the current training dataset when the factor exceeds 4, which is sufficient for reasoning training.
}
\centering
\label{tab:effective len}
\begin{tabular}{@{}c|cccccc@{}}
\toprule
RoPE theta & $\times$1 & $\times$4 & $\times$8 & $\times$16 & $\times$32 & $\times$64 \\ \midrule
Effective Context Length & 9k  & 21k  & 29k  & 32k  & 21k  & 17k  \\
32k NIAH Score           & 0.00  & 3.75 & 58.30 & 77.05 & 58.86 & 35.00 \\
LongBench Score          & 21.14 & 39.21 & 39.78 & 40.41 & 38.96 & 38.01 \\
RULER Score              & 56.13 & 69.57 & 79.62 & 94.24 & 88.07 & 84.98 \\ \bottomrule
\end{tabular}
\end{table}
\begin{figure}[t]
    \centering
    
    \begin{subfigure}{0.32\textwidth}
        \centering
        \includegraphics[width=\linewidth]{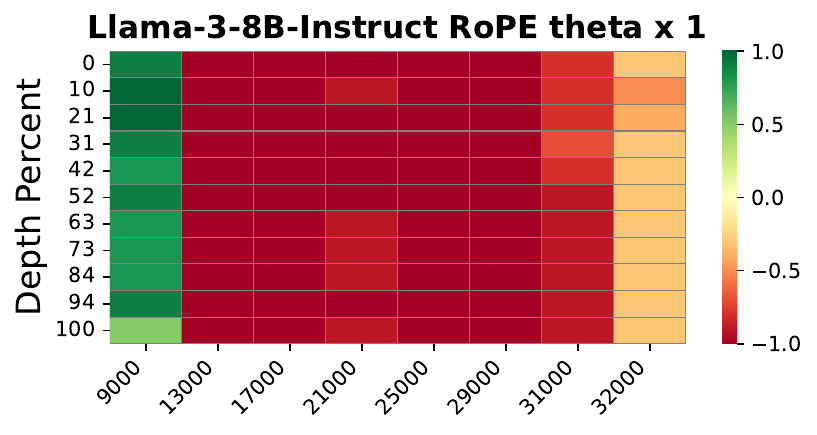}
    \end{subfigure}
    \begin{subfigure}{0.32\textwidth}
        \centering
        \includegraphics[width=\linewidth]{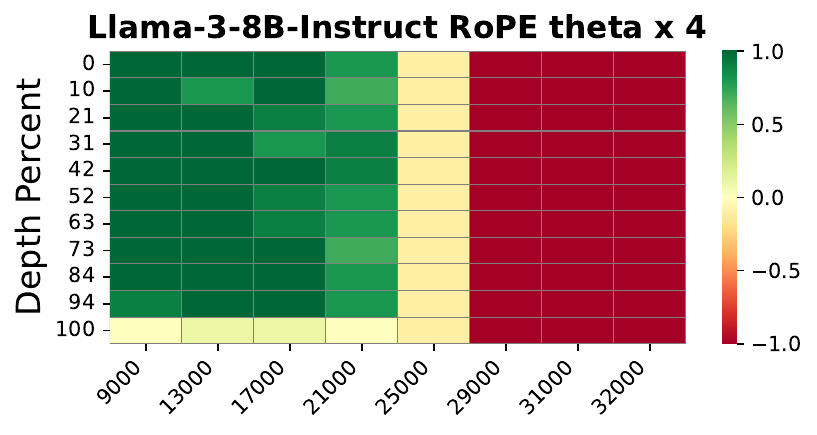}
    \end{subfigure}
    \begin{subfigure}{0.32\textwidth}
        \centering
        \includegraphics[width=\linewidth]{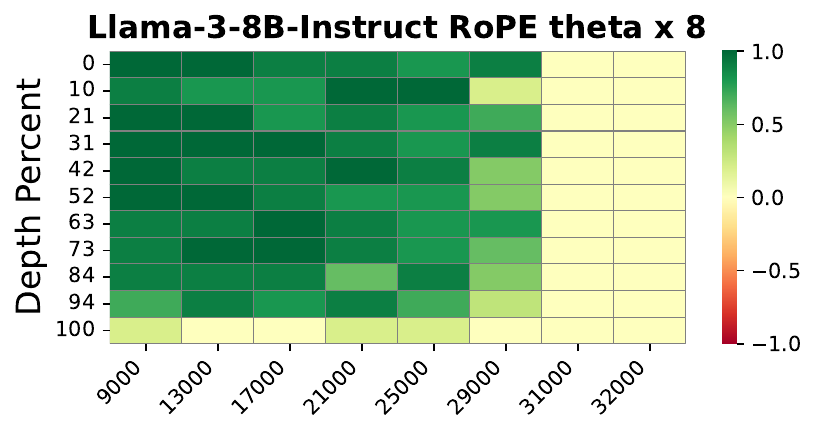}
    \end{subfigure}
    \hfill
    \begin{subfigure}{0.32\textwidth}
        \centering
        \includegraphics[width=\linewidth]{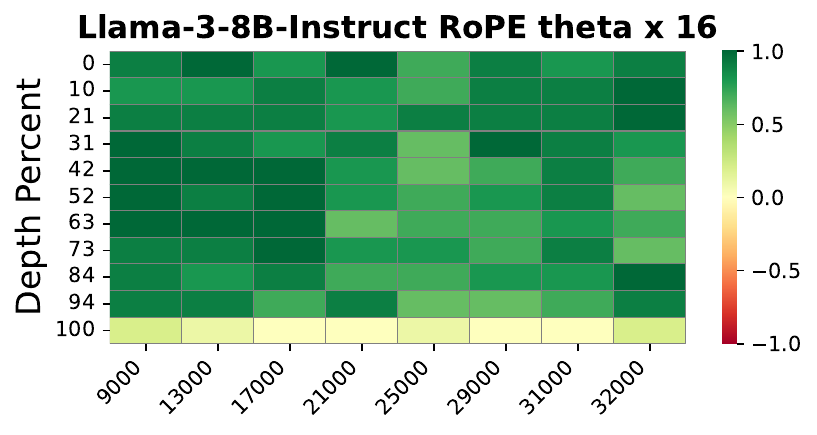}
    \end{subfigure}
    \begin{subfigure}{0.32\textwidth}
        \centering
       \includegraphics[width=\linewidth]{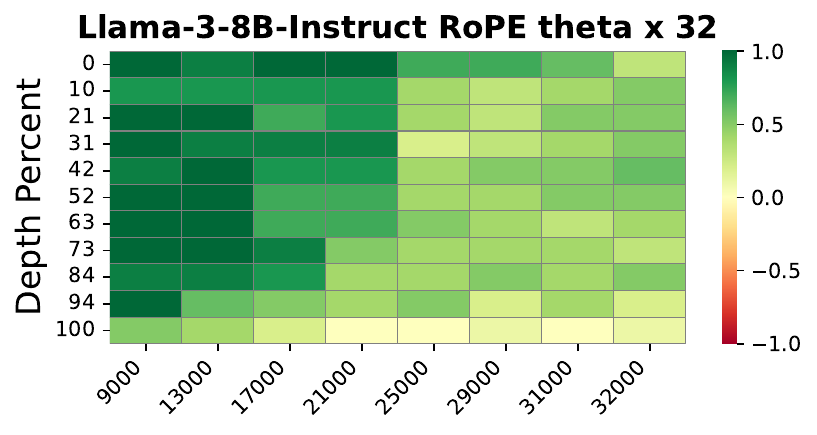}
    \end{subfigure}
    \begin{subfigure}{0.32\textwidth}
        \centering
        \includegraphics[width=\linewidth]{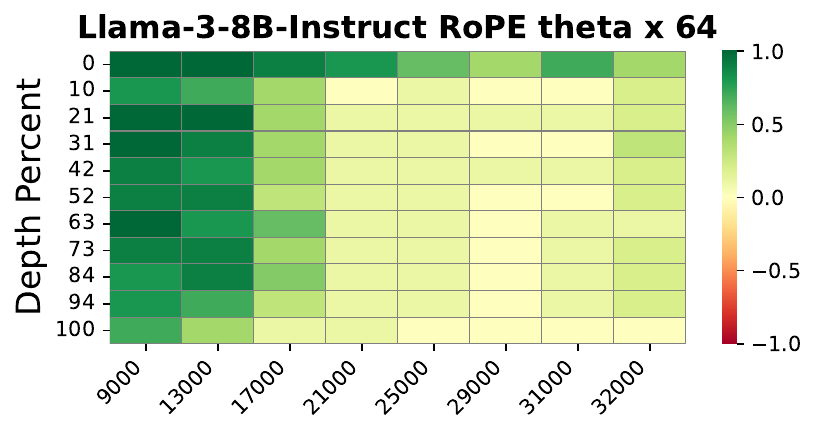}
    \end{subfigure}
    \vspace{-8pt}
    \caption{
    Needle-in-a-Haystack Results for LLaMA-3-8B-Instruct. Performance of LLaMA-3-8B-Instruct on the \textit{Needle-in-a-Haystack} benchmark with a 32K context under different \texttt{RoPE theta} scaling factors. \texttt{RoPE theta x 16} refers to scaling the original \texttt{RoPE theta} by a factor of 16.
    }
    \label{fig:Needle-in-a-Haystack Results for LLaMA-3-8B-Instruct}
\end{figure}
\begin{figure}[t]
    \centering
    \begin{subfigure}{1.0\textwidth}
        \centering
        \includegraphics[width=\linewidth]{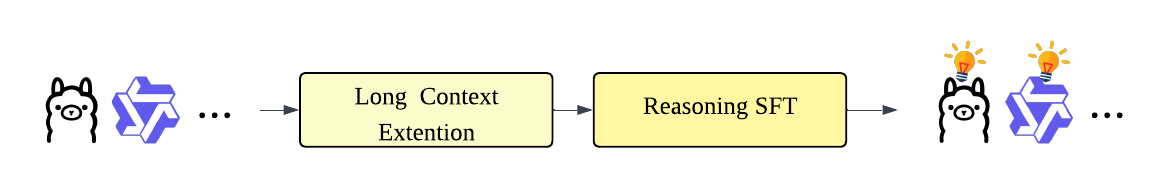}
    \end{subfigure}\vspace{-10pt}
    \caption{
    Pipeline for verifying the connection between reasoning and long-context ability. We first expand the model’s long-context capability to obtain variants with different levels of long-context capacity. Then, we perform SFT on reasoning datasets to obtain reasoning-enhanced models.
    }
    \label{fig:training_pipeline}
\end{figure}

\section{Empirical analysis: verifying the connection between reasoning and
long-Context ability}

In this section, we investigate the connection between a model's long-context capability and its reasoning performance. We begin by applying long-context extension strategies to obtain models with varying levels of long-context ability at 32k tokens. These models are then fine-tuned on different reasoning datasets using supervised fine-tuning (SFT), allowing us to assess how different context capacities influence the effectiveness of reasoning training. Next, we further extend the long-context capability to 128K tokens or beyond, in order to examine whether extreme context lengths can provide additional gains in reasoning performance. Finally, we introduce a recipe for LLM reasoning training: extend the long-context ability before reasoning finetuning and conduct an experiment to verify that this recipe is useful and effective. The overall training pipeline is illustrated in \cref{fig:training_pipeline}.

\subsection{Experimental setup}

\textbf{Long context ability extension strategy.} We use two strategies to enhance long-context capability. The first is directly scaling the \texttt{RoPE theta} parameter by different factors, which has been shown to improve a model’s ability to handle longer sequences~\citep{peng2023ntk}.  The second leverages model merging: we merge the target model with another model that possesses stronger long-context capabilities. In this setting, we carefully control the merge ratio to ensure that the base performance remains nearly unchanged, allowing us to isolate the effect of long-context enhancement as the only influential factor.

\textbf{Data processing for reasoning SFT}. We utilize the OpenR1-Math-220K dataset~\citep{openr1} and divide it into two categories based on response length: \textbf{short} samples (responses within 8K tokens) and \textbf{long} samples (responses ranging from 8K to 16K tokens). For both categories, we sample 20K instances and perform correctness filtering to ensure that each response is factually accurate and correct. These two subsets are then used independently to fine-tune models to improve their reasoning ability.

\textbf{Training details}. All models are fine-tuned using four NVIDIA H200 GPUs. We employ the \texttt{LLaMAFactory} library with a batch size of 32, a learning rate of $1.0 \times 10^{-5}$ and 3 epochs.

\textbf{Long context evaluation}. We adopt the \textit{Needle in a Haystack} benchmark provided by the OpenCompass framework to access long context ability. For simplicity and robustness, we use the accuracy on the single-haystack setting as our primary metric: a correct response receives a score of $+1$, a repetitive/degenerate answer receives $-1$, and an incorrect but non-degenerate response receives $0$. 

\textbf{Reasoning evaluation}. To further evaluate the model’s reasoning ability post-training, we use three math benchmarks: MATH500, AIME22--24, and GSM8K. Following the evaluation methodology from DeepSeek-R1, we adopt the \textit{pass@1(5)} metric, where five responses are generated for each question and accuracy is computed over all the responses. This provides a more stable estimate of reasoning performance and abilities of Large Language Models after finetuning with datasets.

\begin{figure}[ht]
\centering
    
    \begin{subfigure}{0.329\textwidth}
        \centering
        \includegraphics[width=\linewidth]{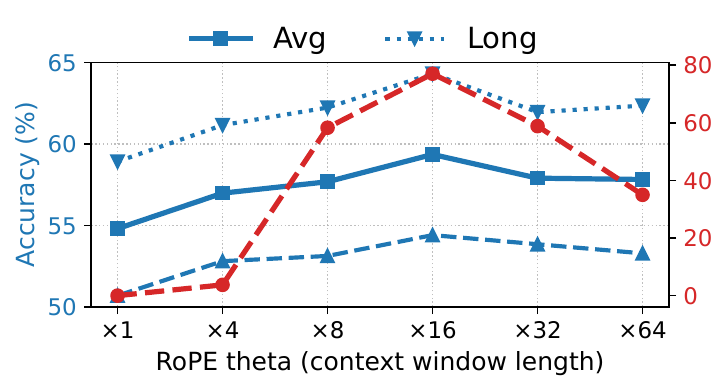}
        \caption{MATH500}
    \end{subfigure}
    \begin{subfigure}{0.329\textwidth}
        \centering
        \includegraphics[width=\linewidth]{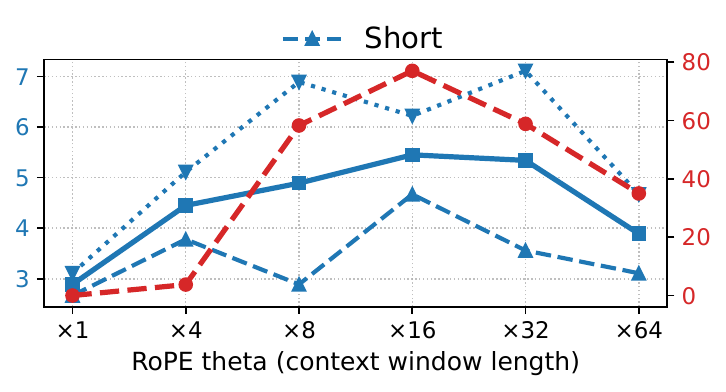}
        \caption{AIME22-24}
    \end{subfigure}
    \begin{subfigure}{0.329\textwidth}
        \centering
        \includegraphics[width=\linewidth]{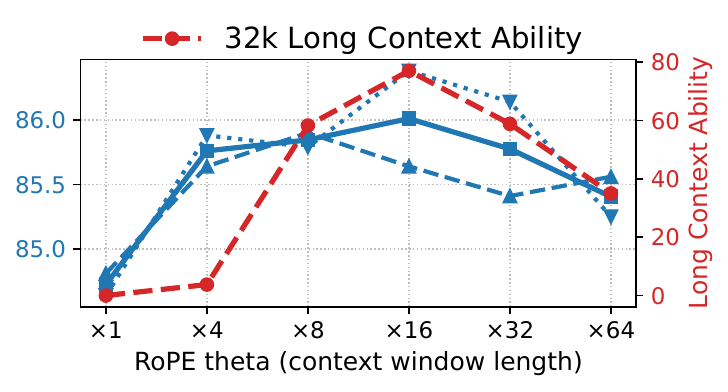}
        \caption{GSM8K}
    \end{subfigure}
    \vspace{-15pt}
    \caption{Visualization of the relationship between 32k long-context ability and reasoning performance across different model variants of \texttt{LLaMA-3.1-8B-Instruct} on three math benchmarks (MATH500, AIME22-24, GSM8K). \texttt{Short} and \texttt{Long} refer to performance after fine-tuning on short and long reasoning datasets. \texttt{Avg} represents the average of the short and long fine-tuned results.}
    \label{fig:Visualization of the relationship llama3}
\end{figure}
\begin{figure}[ht]
    \centering
    \begin{subfigure}{0.32\textwidth}
        \centering
        \includegraphics[width=\linewidth]{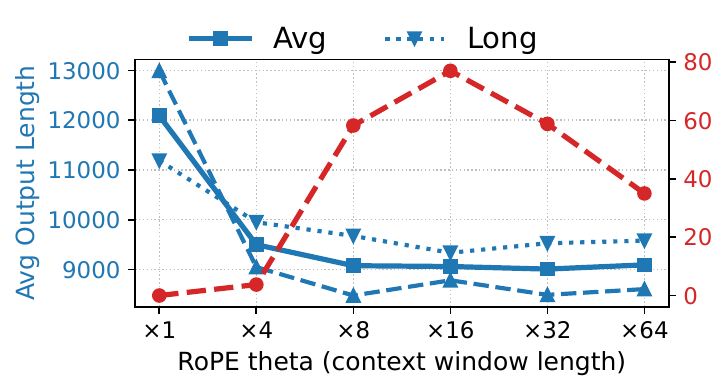}
        \caption{MATH500}
    \end{subfigure}
    \begin{subfigure}{0.32\textwidth}
        \centering
        \includegraphics[width=\linewidth]{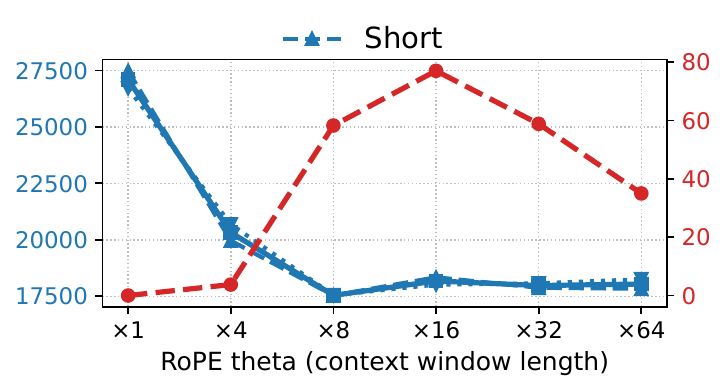}
        \caption{AIME22-24}
    \end{subfigure}
    \begin{subfigure}{0.32\textwidth}
        \centering
        \includegraphics[width=\linewidth]{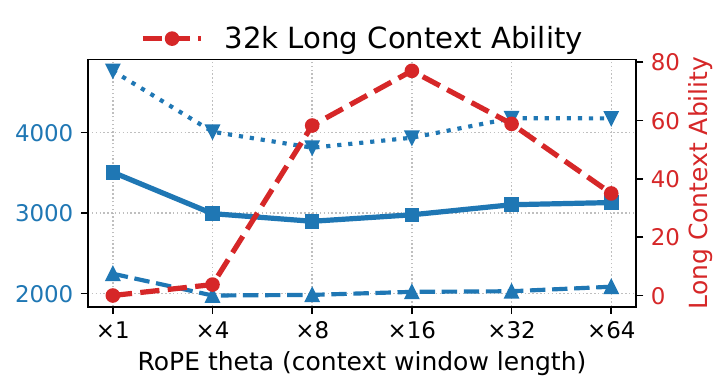}
        \caption{GSM8K}
    \end{subfigure}
    \vspace{-5pt}
    \caption{Visualization of the relationship between 32k long-context ability and average output length across different model variants of \texttt{LLaMA-3.1-8B-Instruct} on MATH500, AIME22-24, GSM8K.}
    \label{fig: Visualization of the relationship llama3 2}
\end{figure}

\subsection{How does long-context capacity affect reasoning SFT?}
\label{sec:appendix_outputlength}

In this experiment, we investigate how different levels of long-context capability affect the performance of reasoning SFT (Supervised Finetuing). We use \texttt{LLaMA3-8B-Instruct} as the base model, which originally supports up to 8K tokens. 
% To enable long-context training, we extend the context window to 32K tokens. 
To obtain models with varying long-context capabilities, we scale the \texttt{RoPE} \texttt{theta} value by the following different factors: \{1, 4, 8, 16, 32, 64\}.

First, we evaluate the long-context ability of each variant using the \textit{Needle-in-a-Haystack} benchmark. As shown in \cref{fig:Needle-in-a-Haystack Results for LLaMA-3-8B-Instruct} and \cref{tab:effective len}, performance improves as the \texttt{RoPE theta} scaling factor increases from $1$ to $16$, reaching peak performance at $16$. Beyond this point, further increases in \texttt{theta} lead to performance degradation. What's more, when the scaling theta is more than 4, the model's effective length is more than 16k, which is enough for the reasoning training for current datasets.

Next, we perform supervised fine-tuning (SFT) using both \texttt{short} and \texttt{long} reasoning datasets on models with different \texttt{RoPE theta} settings. The detailed results are summarized in \cref{tab:Effect of RoPE theta Scaling}. In addition, we plot the relationship between each model's long-context capability and its reasoning performance. We also visualize how long-context ability correlates with the average output length. The results are shown in \cref{fig:Visualization of the relationship llama3,fig: Visualization of the relationship llama3 2}. We observe that the best reasoning performance across all three evaluation benchmarks consistently occurs when the \texttt{RoPE theta} scaling factor is set to 16—coinciding with the strongest long-context capability observed in the Needle-in-a-Haystack task. In contrast, models with suboptimal \texttt{theta} settings exhibit reduced accuracy undergoing the same SFT procedure. These findings suggest that enhancing long-context processing ability can directly contribute to improved reasoning performance after reasoning SFT (Supervised Fine-tuning). 

Furthermore, we observe that, in general, models fine-tuned on \texttt{long} reasoning datasets outperform those fine-tuned on \texttt{short} datasets in both \texttt{MATH500} and \texttt{AIME}. These long-form datasets typically contain more complex problems and richer intermediate reasoning steps, making them more effective for improving the model’s reasoning ability. However, in order to benefit from such data, models must possess sufficiently strong and effective long-context capabilities to process the extended inputs.

\begin{table}[t]
\caption{
\textbf{Effect of RoPE theta scaling on long-context and reasoning performance on MATH500.} 
Evaluation of \texttt{LLaMA3-8B-Instruct} with different \texttt{RoPE theta} scaling factors on the 32K \textit{Needle-in-a-Haystack} benchmark. 
\texttt{Base} refers to the model's performance before SFT, while \texttt{Short} and \texttt{Long} denote results after fine-tuning on short and long reasoning datasets, respectively. \texttt{Avg} represents the average of the \texttt{Short} and \texttt{Long} performances.
}
\label{tab:Effect of RoPE theta Scaling}
\small{
\resizebox{\textwidth}{!}{
\begin{tabular}{@{}c|c|cccccccc@{}}
\toprule
RoPE       & 32k long       & \multicolumn{4}{c}{Acc(\%)}                                                            & \multicolumn{4}{c}{Average Output Length}            \\ \cmidrule(l){3-10} 
theta      & ctx ability    & Base           & Short          & Long           & \multicolumn{1}{c|}{Avg}            & Base & Short         & Long          & Avg           \\ \midrule
$\times1$  & 0              & \textbf{24.40} & 50.68          & 58.92          & \multicolumn{1}{c|}{\cellcolor{gray!20}{54.80}}          & 746  & 12991         & 11187         & \cellcolor{gray!20}{12089}         \\
$\times4$  & 3.75           & 20.40          & 52.80          & 61.16          & \multicolumn{1}{c|}{\cellcolor{gray!20}{56.98}}          & 754  & 9047          & 9949          & \cellcolor{gray!20}{9498}          \\
$\times8$  & 58.30          & 18.96          & 53.12          & 62.24          & \multicolumn{1}{c|}{\cellcolor{gray!20}{57.68}}          & 767  & \textbf{8476} & 9673          & \cellcolor{gray!20}{9075}          \\
$\times16$ & \textbf{77.05} & 17.28          & \textbf{54.40} & \textbf{64.32} & \multicolumn{1}{c|}{\cellcolor{gray!20}{\textbf{59.36}}} & 608  & 8782          & \textbf{9335} & \cellcolor{gray!20}{9059} \\
$\times32$ & 58.86          & 15.24          & 53.84          & 61.96          & \multicolumn{1}{c|}{\cellcolor{gray!20}{57.90}}          & 490  & 8489          & 9526          & \cellcolor{gray!20}{\textbf{9007}}         \\
$\times64$ & 35.00          & 14.20          & 53.28          & 62.36          & \multicolumn{1}{c|}{\cellcolor{gray!20}{57.82}}          & 453  & 8605          & 9579          & \cellcolor{gray!20}{9092}         \\ \bottomrule
\end{tabular}
}
}
\end{table}

\subsection{Generality and robustness of long-context gains to reasoning ability }
\label{sec:generality}

We investigate whether the benefits of stronger long-context ability generalize across tasks, model families, and input lengths. Our results show that the improvements are not limited to math reasoning. 

\textbf{Cross-domain generalization.} We fine-tuned \texttt{LLaMA3-8B-Instruct} models with RoPE scales $\times$1, $\times$4, and $\times$16 on 20k samples from the \textit{science} and \textit{code} domains of \texttt{OpenThoughts3-1.2M}.  
Evaluations on GPQA and Livecode (\Cref{tab:generalization-science-code}) show that stronger long-context ability consistently improves accuracy beyond math reasoning.

\begin{table}[t]
\centering
\caption{Cross-domain evaluation of long-context ability on GPQA (science) and Livecode (code).}
\label{tab:generalization-science-code}
\begin{tabular}{l|c|ccc|ccc}
\toprule
Task & Setting & \multicolumn{3}{c|}{GPQA Accuracy (\%)} & \multicolumn{3}{c}{Livecode Accuracy (\%)} \\
\midrule
 &  & $\times$1 & $\times$4 & $\times$16 & $\times$1 & $\times$4 & $\times$16 \\
\midrule
32k NIAH & -      & 0.00  & 3.75  & 77.05 & 0.00  & 3.75  & 77.05 \\
Accuracy & before training & 31.82 & 31.31 & 30.81 & 12.30 & 10.16 & 7.45  \\
         & after training& 37.27 & 39.19 & 41.92 & 25.34 & 27.95 & 32.27 \\
\bottomrule
\end{tabular}
\end{table}

\textbf{Across model families.} We trained \texttt{Phi-4} (14B) with the same RoPE settings. Consistent gains on MATH500 (\Cref{tab:phi4-math500}) confirm that the benefit is not tied to a specific architecture or scale. The results of AIME22-24 are in \Cref{tab:phi4-aime} and it has the similar performance like MATH500.

\begin{table}[t]
\centering
\caption{Performance of \texttt{Phi-4} under different RoPE scales on MATH500.}
\label{tab:phi4-math500}
\begin{tabular}{l|c|c|c|c}
\toprule
Model & RoPE & 32k NIAH (\%) & MATH500 (before, \%) & MATH500 (after, \%) \\
\midrule
Phi-4 & $\times$1  & 52.27 & 79.52 & 88.62 \\
Phi-4 & $\times$4  & 78.07 & 77.48 & 89.14 \\
Phi-4 & $\times$16 & 84.77 & 73.20 & 89.90 \\
\bottomrule
\end{tabular}
\end{table}

\textbf{Short-input reasoning.} On the short-input \texttt{MMLU-STEM} benchmark, models with stronger long-context ability also achieve higher post-training accuracy (\Cref{tab:mmlu-short}). This indicates that long-context training not only preserves but can reinforce reasoning on short-input tasks.

\begin{table}[t]
\centering
\caption{Effect of long-context ability on short-input reasoning (MMLU-STEM) with \texttt{LLaMA3-8B}.}
\label{tab:mmlu-short}
\begin{tabular}{c|c|c|c}
\toprule
RoPE Scale & 32k NIAH (\%) & MMLU-STEM (before, \%) & MMLU-STEM (after, \%) \\
\midrule
$\times$1  & 0.00  & 54.36 & 71.06 \\
$\times$4  & 3.75  & 53.85 & 73.04 \\
$\times$16 & 77.05 & 51.44 & 74.27 \\
\bottomrule
\end{tabular}
\end{table}

\medskip
Overall, these results demonstrate that the gains from stronger long-context ability are robust: they extend beyond math to science and code, hold across LLaMA, Qwen, and Phi model families, and even benefit short-input reasoning.

\subsection{Does extremely long context bring further gains on reasoning SFT?}

In previous experiments, we extended models to handle up to 32K tokens and observed notable improvements in reasoning performance. A natural question arises: \textit{can even stronger long-context capabilities further enhance reasoning}, or is there a limit beyond which performance saturates or even degrades? To explore this, we conduct experiments using models with a context length of 1M tokens (Qwen2.5-7B-Instruct-1M)—far beyond the typical range of existing reasoning datasets.

We adopt a linear merging strategy to construct models with varying long-context capacities while minimizing changes to their base ability. Specifically, we merge two models with different context capabilities at various ratios to obtain models with intermediate long-context strengths. With selected merge ratio, we ensure that the base capabilities remain largely unchanged, isolating long-context capacity as the key variable. We apply this strategy to two models: Qwen2.5-7B-Instruct-1M with long context ability at 1M and Qwen2.5-7B-Instruct with long context ability at 32k. 

We first evaluate the merged models on \textit{Needle-in-a-Haystack} at 32k and find that long-context ability grows with the merge ratio, but near-perfect scores make it less discriminative (\cref{app:Other Results on Different Long-Context Ability}). We therefore adopt more challenging 32k tasks such as \textit{Value Tracking} and \textit{Question Answering}, which better capture effective long-context processing by requiring stronger long-range reasoning.

\begin{figure}[t]
\centering
    \begin{subfigure}{0.32\textwidth}
        \centering
        \includegraphics[width=\linewidth]{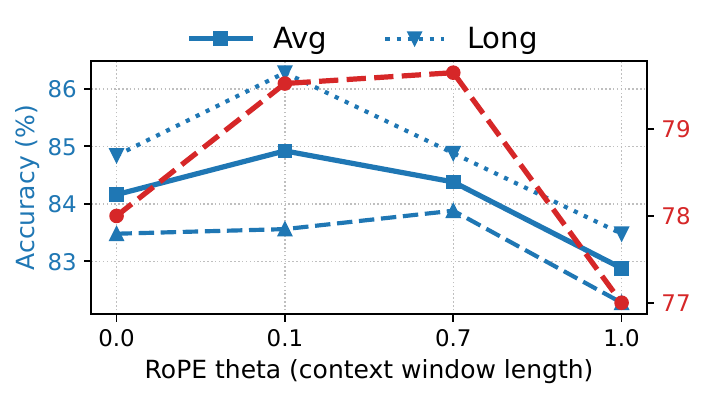}
        \caption{MATH500}
    \end{subfigure}
    \begin{subfigure}{0.32\textwidth}
        \centering
        \includegraphics[width=\linewidth]{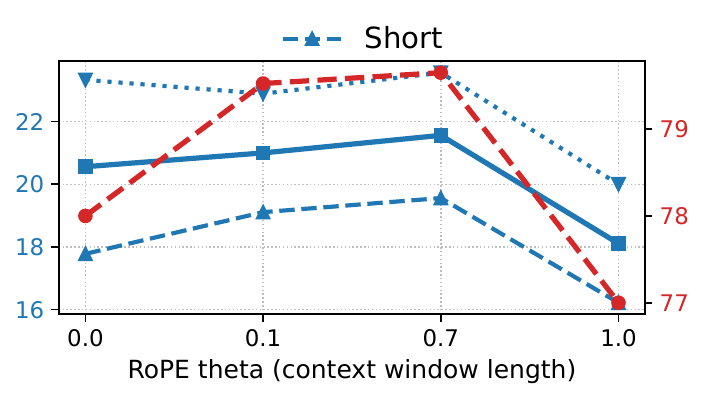}
        \caption{AIME22-24}
    \end{subfigure}
    \begin{subfigure}{0.32\textwidth}
        \centering
        \includegraphics[width=\linewidth]{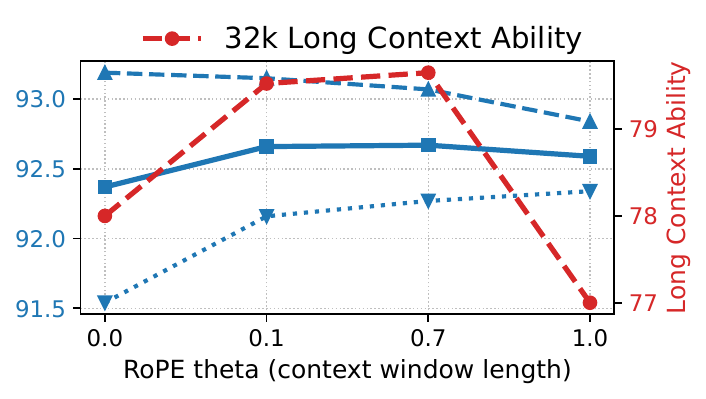}
        \caption{GSM8K}
    \end{subfigure}
    \vspace{-5pt}
    \caption{Visualization of the relationship between 32k long-context ability and reasoning performance across different model variants of \texttt{Qwen2.5-7B-Instruct} on MATH500, AIME22-24, GSM8K. A merge ratio of 0.1 indicates that the long-context variant (\texttt{Qwen2.5-7B-Instruct-1M}) contributes 10\% to the final merged model. \texttt{Short} and \texttt{Long} refer to performance after fine-tuning on short and long reasoning datasets. \texttt{Avg} represents the average of the short and long fine-tuned results.}
    \label{fig:Visualization of the relationship between 128k long-context ability}
\end{figure}

\begin{figure}[t]
\centering
    \begin{subfigure}{0.32\textwidth}
        \centering
        \includegraphics[width=\linewidth]{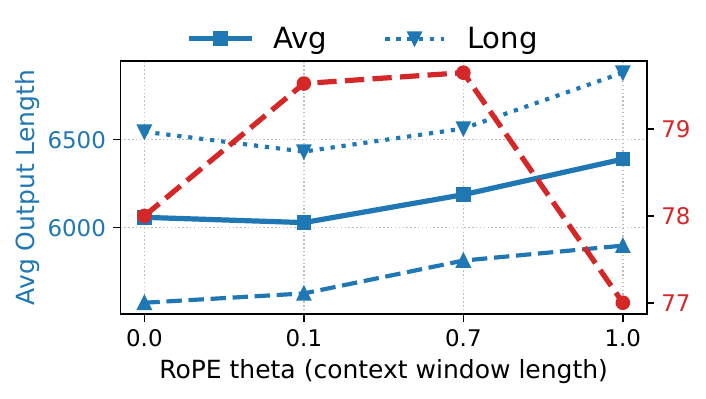}
        \caption{MATH500}
    \end{subfigure}
    \begin{subfigure}{0.32\textwidth}
        \centering
        \includegraphics[width=\linewidth]{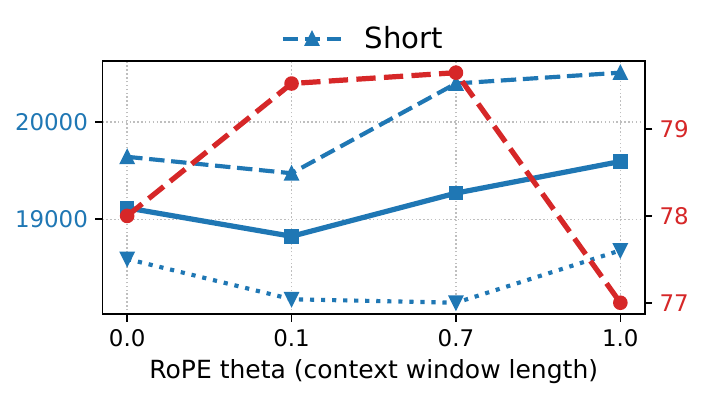}
        \caption{AIME22-24}
    \end{subfigure}
    \begin{subfigure}{0.32\textwidth}
        \centering
        \includegraphics[width=\linewidth]{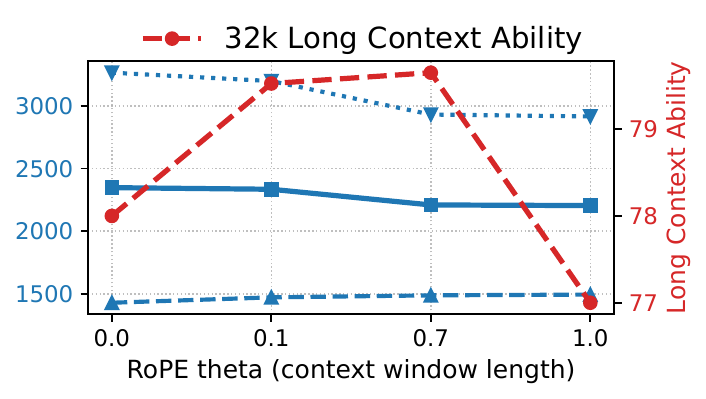}
        \caption{GSM8K}
    \end{subfigure}
    \vspace{-5pt}
    \caption{The relationship between 32k long-context ability and average output length across different model variants of \texttt{Qwen2.5-7B-Instruct} on three math benchmarks.
    } 
    \label{fig:Visualization of the relationship between 128k long-context ability2}
\end{figure}

We fine-tune the merged models on both \texttt{short} and \texttt{long} reasoning datasets, and evaluate on \texttt{MATH500}, \texttt{AIME22-24}, and \texttt{GSM8K}. We also examine the relationship between effective long-context ability, reasoning accuracy, and output length (\cref{fig:Visualization of the relationship between 128k long-context ability,fig:Visualization of the relationship between 128k long-context ability2}). Results show that moderate merge ratios (e.g., 0.1, 0.7) yield strong effective long-context ability and high reasoning accuracy, while the 1M model (ratio 1.0) exhibits weaker effective long-context utilization and degraded performance.

\begin{table}[t]
\caption{
\textbf{Validation of the proposed fine-tuning recipe for reasoning.} 
We evaluate the effectiveness of our proposed training recipe using \texttt{Qwen2.5-Math-7B-Instruct}, a model with an initial long-context capacity limited to 4K tokens. We first scale its \texttt{RoPE theta} by a factor of 16, and then merge it with \texttt{Qwen2.5-7B-Instruct-1M} using a merge ratio of $0.3$. We observe consistent improvement in reasoning performance after each step, demonstrating the effectiveness of enhancing long-context capability prior to reasoning fine-tuning to get a higher reasoning ability.
}
\label{tab:Validation of the Proposed Fine-Tuning Recipe for Reasoning}
\resizebox{\textwidth}{!}{
\begin{tabular}{@{}c|cccc|cccc@{}}
\toprule
\multirow{2}{*}{Operation}      & \multicolumn{4}{c|}{Acc(\%)}                                                           & \multicolumn{4}{c}{Avg Output Length}                    \\ \cmidrule(l){2-9} 
                                & Base           & Short          & Long           & Avg            & Base  & Short          & Long           & Avg            \\ \midrule
\multicolumn{9}{c}{MATH500}                                                                                                                                                                                                                   \\ \midrule
RoPE theta $\times1$                   & \textbf{81.88} & 86.28          & 83.80          & \cellcolor{gray!20}{85.04}          & 2037  & 1022           & 2147           & \cellcolor{gray!20}{1584}           \\
RoPE theta $\times16$                  & 65.16          & 87.68          & 88.72          & \cellcolor{gray!20}{88.20}          & 4579  & 4213           & 5789           & \cellcolor{gray!20}{5501}         \\
0.7 $\times$ RoPE theta $\times16$ + 0.3 $\times$  1M-Model & 74.12          & \textbf{88.28} & \textbf{89.12} & \cellcolor{gray!20}{\textbf{88.70}} & 2228  & \textbf{4948}  & \textbf{6515}  & \cellcolor{gray!20}{\textbf{5731}}  \\ \midrule
\multicolumn{9}{c}{AIME22-24}                                                                                                                                                                                                                 \\ \midrule
RoPE theta $\times1$                   & \textbf{8.44}  & 16.22          & 13.78          & \cellcolor{gray!20}{15.00}          & 8257  & 3463           & 10418          & \cellcolor{gray!20}{6940}          \\
RoPE theta $\times16$                  & 3.78           & 25.78          & 27.56          & \cellcolor{gray!20}{26.67}          & 13411 & 15036          & 17321          & \cellcolor{gray!20}{16179}         \\
0.7 $\times$ RoPE theta $\times16$ + 0.3 $\times$ 1M-Model & 7.11           & \textbf{26.67} & \textbf{29.33} & \cellcolor{gray!20}{\textbf{28.00}} & 6859  & \textbf{20265} & \textbf{22919} & \cellcolor{gray!20}{\textbf{21592}} \\ \bottomrule
\end{tabular}
}
\end{table}

\begin{figure}[t]
    \centering
    \begin{subfigure}{0.32\textwidth}
        \centering
        \includegraphics[width=\linewidth]{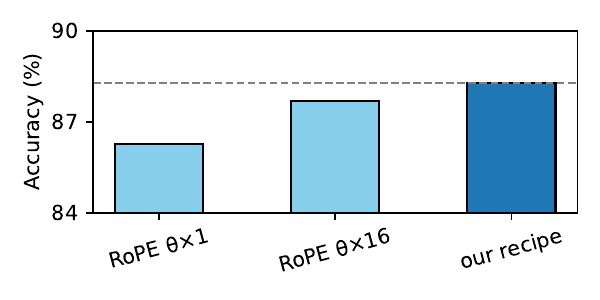}
        \caption{MATH500 (Short)}
        \label{fig:math500_short}
    \end{subfigure}
    \hfill
    \begin{subfigure}{0.32\textwidth}
        \centering
        \includegraphics[width=\linewidth]{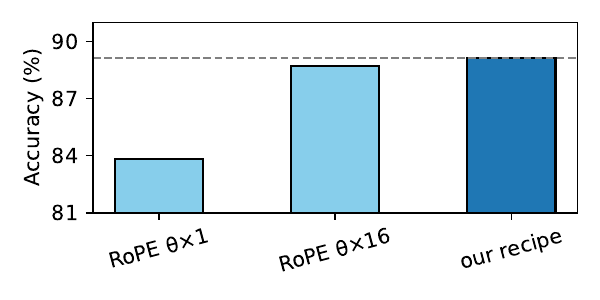}
        \caption{MATH500 (Long)}
        \label{fig:math500_long}
    \end{subfigure}
    \hfill
    \begin{subfigure}{0.32\textwidth}
        \centering
        \includegraphics[width=\linewidth]{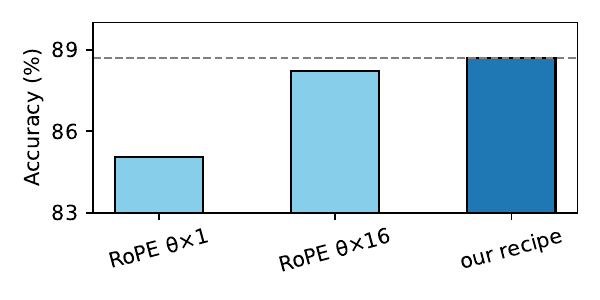}
        \caption{MATH500 (Avg)}
        \label{fig:math500_avg}
    \end{subfigure}

    \vspace{0.4em} 
    \begin{subfigure}{0.32\textwidth}
        \centering
        \includegraphics[width=\linewidth]{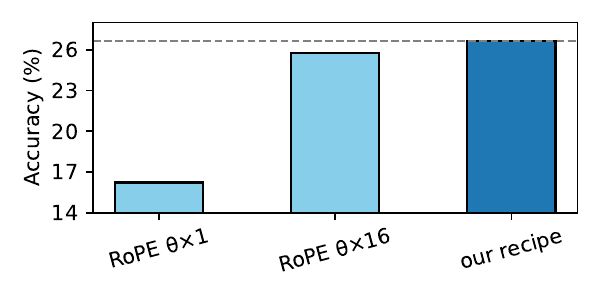}
        \caption{AIME22-24 (Short)}
        \label{fig:aime_short}
    \end{subfigure}
    \hfill
    \begin{subfigure}{0.32\textwidth}
        \centering
        \includegraphics[width=\linewidth]{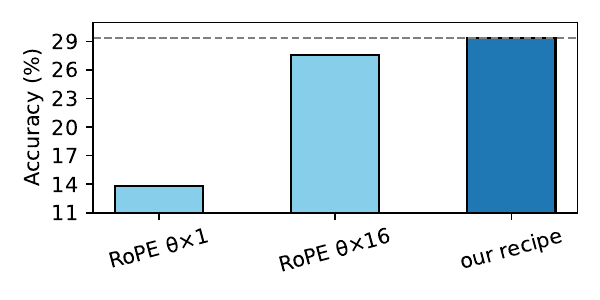}
        \caption{AIME22-24 (Long)}
        \label{fig:aime_long}
    \end{subfigure}
    \hfill
    \begin{subfigure}{0.32\textwidth}
        \centering
        \includegraphics[width=\linewidth]{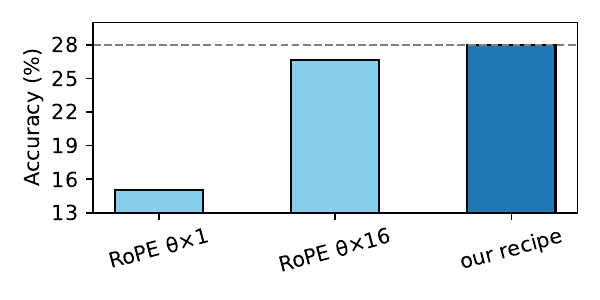}
        \caption{AIME22-24 (Avg)}
        \label{fig:aime_avg}
    \end{subfigure}
    \caption{
    Comparison of reasoning accuracy across different configurations.
    Results are reported on \texttt{MATH500} (top row) and \texttt{AIME22-24} (bottom row), with separate evaluation for short, long, and average context cases. 
    Our proposed recipe (dark blue bar, dashed line) consistently outperforms the baseline \texttt{RoPE theta $\times1$ } and the extended-context baseline \texttt{RoPE theta $\times16$ }, demonstrating that enhancing long-context capability prior to reasoning fine-tuning yields stable and significant accuracy improvements, enhancing models' reasoning ability. 
    }
\end{figure}

\subsection{Proposed reasoning recipe: extend context length first}

Based on our previous experiments, we propose a training \textbf{recipe} for improving reasoning capabilities via supervised fine-tuning (SFT): \textit{first appropriately enhance the model's long-context capability, then apply reasoning-specific SFT}. This recipe aims to better prepare the model for long-form reasoning tasks, where both context length and reasoning complexity are critical. To validate this approach, we experiment with the \texttt{Qwen2.5-Math-7B-Instruct} model, which demonstrates strong mathematical performances but has a limited context length of 4k tokens, which is a good example of the recipe.

We enhance its long-context capability by multiplying the \texttt{RoPE theta} value by a scaling factor of $16$ and merging the model with \texttt{Qwen2.5-7B-Instruct-1M} using a merge ratio of $0.3$. After the merging step, the long-context ability improves noticeably, as is shown in \cref{app:Other Results on Different Long-Context Ability}. Subsequently, we perform SFT using both \texttt{short} and \texttt{long} reasoning datasets, and evaluate the fine-tuned models on two benchmarks: \texttt{MATH500} and \texttt{AIME22-24}. As shown in \cref{tab:Validation of the Proposed Fine-Tuning Recipe for Reasoning}, across both short and long reasoning datasets, the model variant with \texttt{RoPE theta} scaled by 16 and a merge ratio of $0.3$ consistently achieves the highest performance after reasoning SFT on short and long datasets.

\section{Related Works}

\textbf{Reasoning models.}  
The release of \texttt{DeepSeek-R1}~\citep{deepseekai2025deepseekr1incentivizingreasoningcapability} has catalyzed widespread interest in open-source reasoning models~\citep{zhang2024llm, plaat2024reasoning, ke2025survey}. 
Most existing approaches to improving reasoning rely on supervised fine-tuning (SFT) and rule-based reinforcement learning (RL)~\citep{shao2024deepseekmathpushinglimitsmathematical, zhong2024dpo, schulman2017proximalpolicyoptimizationalgorithms, yu2025dapoopensourcellmreinforcement}. 
For instance, \texttt{SkyThought}~\citep{sky_t1_2025} and \texttt{Bespoke}~\citep{bespoke_stratos} fine-tune models using responses generated by larger reasoning models such as \texttt{QwQ-32B}~\citep{qwq32b} and \texttt{DeepSeek-R1}. 
Other methods like \texttt{S1}~\citep{muennighoff2025s1simpletesttimescaling} and \texttt{LIMO}~\citep{ye2025limoreasoning} highlight the importance of using compact yet high-quality reasoning samples to improve data efficiency. These efforts have led to the creation of numerous open-source reasoning datasets, including \texttt{OpenR1-Math-220K}~\citep{openr1}, \texttt{DeepMath-103K}~\citep{deepmath}, \texttt{OpenThoughts}~\citep{openthoughts}, and others~\citep{moshkov2025aimo2, slam-distillation-from-r1, bercovich2025llamanemotronefficientreasoningmodels}. 
Notable models trained on these datasets include \texttt{OpenR1-Qwen-7B} and \texttt{DeepMath-Zero-7B}. 
Subsequent work~\citep{li2025llmseasilylearnreason, ji2025first} has proposed techniques to further optimize SFT and RL pipelines, arguing that structural aspects of reasoning data (e.g., step-wise decomposition) can be more impactful than content alone.  More recent studies~\citep{abdin2025phi4reasoningtechnicalreport, yao2025optimizingchainofthoughtreasonersgradient, ni2025teachinglargelanguagemodels, zeng2025simplerlzooinvestigatingtamingzero} aim to improve the alignment and generalization of reasoning through better data construction, gradient control, and training optimization strategies. Despite these advances, an important factor remains underexplored: the role of \texttt{long-context capability} in reasoning. While recent datasets and tasks increasingly involve longer sequences, no existing work systematically investigates how a model’s ability to process extended contexts affects its reasoning performance. Our work aims to figure out the relationship between them.

\textbf{Long-context ability.} 
Closed-source models such as \texttt{GPT-4}~\citep{achiam2023gpt}, \texttt{Claude}~\citep{caruccio2024claude}, and Gemini~\citep{team2024gemini} already support context lengths of 128K tokens or more, enabled by advances in both pre-training and post-training~\citep{dao2022flashattention,xiong2023effective,hsu2024liger,li2021sequence}.  
In parallel, open-source models—including \texttt{LLaMA}~\citep{grattafiori2024llama3herdmodels}, \texttt{Qwen}~\citep{qwen2.5,yang2025qwen3technicalreport}, and \texttt{Phi}~\citep{abdin2024phi4technicalreport}—have also extended their capacities, with some reaching 1M tokens, such as \texttt{Qwen-2.5-7B-Instruct-1M}~\citep{yang2025qwen251mtechnicalreport} and \texttt{LLaMA-3.1-Nemotron-8B-UltraLong-4M-Instruct}~\citep{ulralong2025}. To enable long context, several approaches have been proposed. \emph{RoPE-based methods} adapt positional encoding for extrapolation without full retraining, including Position Interpolation~\citep{chen2023extending}, NTK Scaling~\citep{peng2023ntk}, YaRN~\citep{peng2023yarn}, and SelfExtend~\citep{jin2024llm}.  
\emph{Attention redesigns} improve scalability or memory retention, such as StreamingLLM~\citep{xiao2023efficient}, LM-Infinite~\citep{han2024lm}, Inf-LLM~\citep{xiao2024infllm}, and Landmark Attention~\citep{mohtashami2023landmark}.  
Another direction is \emph{input compression}, which reduces sequence length by summarization or filtering~\citep{jiang2023longllmlingua,li2023compressing}.  
Finally, recent work explores hardware-aligned designs, including sparse attention architectures~\citep{yuan2025nativesparseattentionhardwarealigned} and MoBA~\citep{lu2025mobamixtureblockattention}.

\section{Discussion and Conclusions}

\textbf{Limitations}: Our analysis is limited to 7B–8B models and does not cover larger scales. We also focus on supervised fine-tuning, leaving open how reinforcement learning interacts with long-context ability. Future work should examine these directions across model families and scales.

This paper investigates the overlooked yet crucial role of long-context capability in reasoning performance. Through behavioral analysis and controlled experiments, we demonstrate a consistent correlation between long-context and downstream reasoning ability. Our results show that models with improved long-context capacity not only perform better on tasks involving lengthy inputs but also achieve higher reasoning accuracy even on short-form benchmarks. Furthermore, we find that enhancing long-context ability prior to supervised fine-tuning yields significant gains across multiple reasoning datasets. Based on these findings, we advocate for a training recipe: first, extend the model's long-context ability, then apply reasoning-specific fine-tuning.

\clearpage
\section*{Acknowledgments}
% This research was partially supported by NSF Awards OAC-2117439. Further, this work made use of the High Performance Computing Resource in the Core Facility for Advanced Research Computing at Case Western Reserve University (CWRU). We give special thanks to the CWRU HPC team for their prompt and professional help and maintenance. The views and conclusions in this paper are those of the authors and do not represent the views of any funding or supporting agencies.

This work was supported in part by NSF grants 2112606 and 2117439. Further, this research made use of the High Performance Computing Resource in the Core Facility for Advanced Research Computing at Case Western Reserve University (CWRU). We give special thanks to the CWRU HPC team for their prompt and professional help and maintenance.

\bibliographystyle{unsrt}
\bibliography{ref}

@article{ke2025survey,
  title={A Survey of Frontiers in LLM Reasoning: Inference Scaling, Learning to Reason, and Agentic Systems},
  author={Ke, Zixuan and Jiao, Fangkai and Ming, Yifei and Nguyen, Xuan-Phi and Xu, Austin and Long, Do Xuan and Li, Minzhi and Qin, Chengwei and Wang, Peifeng and Savarese, Silvio and others},
  journal={arXiv preprint arXiv:2504.09037},
  year={2025}
}

@misc{slam-distillation-from-r1,  
    author = {Sathwik Tejaswi Madhusudhan and Shruthan Radhakrishna and Jash Mehta and Toby Liang},  
    title = {Millions scale dataset distilled from R1-32b},  
    howpublished = {https://huggingface.co/datasets/ServiceNow-AI/R1-Distill-SFT},
    publisher = {SLAM - ServiceNow Language Models Lab}, 
    year = {2025}
}

@misc{bercovich2025llamanemotronefficientreasoningmodels,
      title={Llama-Nemotron: Efficient Reasoning Models}, 
      author={Akhiad Bercovich and Itay Levy and Izik Golan and Mohammad Dabbah and Ran El-Yaniv and Omri Puny and Ido Galil and Zach Moshe and Tomer Ronen and Najeeb Nabwani and Ido Shahaf and Oren Tropp and Ehud Karpas..},
      year={2025},
      eprint={2505.00949},
      archivePrefix={arXiv},
      primaryClass={cs.CL},
      url={https://arxiv.org/abs/2505.00949}, 
}

@article{caruccio2024claude,
  title={Claude 2.0 large language model: Tackling a real-world classification problem with a new iterative prompt engineering approach},
  author={Caruccio, Loredana and Cirillo, Stefano and Polese, Giuseppe and Solimando, Giandomenico and Sundaramurthy, Shanmugam and Tortora, Genoveffa},
  journal={Intelligent Systems with Applications},
  volume={21},
  pages={200336},
  year={2024},
  publisher={Elsevier}
}

@article{li2023compressing,
  title={Compressing context to enhance inference efficiency of large language models},
  author={Li, Yucheng and Dong, Bo and Lin, Chenghua and Guerin, Frank},
  journal={arXiv preprint arXiv:2310.06201},
  year={2023}
}

@article{jiang2023longllmlingua,
  title={Longllmlingua: Accelerating and enhancing llms in long context scenarios via prompt compression},
  author={Jiang, Huiqiang and Wu, Qianhui and Luo, Xufang and Li, Dongsheng and Lin, Chin-Yew and Yang, Yuqing and Qiu, Lili},
  journal={arXiv preprint arXiv:2310.06839},
  year={2023}
}

@article{xiong2023effective,
  title={Effective long-context scaling of foundation models},
  author={Xiong, Wenhan and Liu, Jingyu and Molybog, Igor and Zhang, Hejia and Bhargava, Prajjwal and Hou, Rui and Martin, Louis and Rungta, Rashi and Sankararaman, Karthik Abinav and Oguz, Barlas and others},
  journal={arXiv preprint arXiv:2309.16039},
  year={2023}
}

@article{xiao2023efficient,
  title={Efficient streaming language models with attention sinks},
  author={Xiao, Guangxuan and Tian, Yuandong and Chen, Beidi and Han, Song and Lewis, Mike},
  journal={arXiv preprint arXiv:2309.17453},
  year={2023}
}

@article{li2021sequence,
  title={Sequence parallelism: Long sequence training from system perspective},
  author={Li, Shenggui and Xue, Fuzhao and Baranwal, Chaitanya and Li, Yongbin and You, Yang},
  journal={arXiv preprint arXiv:2105.13120},
  year={2021}
}

@article{dao2022flashattention,
  title={Flashattention: Fast and memory-efficient exact attention with io-awareness},
  author={Dao, Tri and Fu, Dan and Ermon, Stefano and Rudra, Atri and R{\'e}, Christopher},
  journal={Advances in Neural Information Processing Systems},
  volume={35},
  pages={16344--16359},
  year={2022}
}

@article{hsu2024liger,
  title={Liger Kernel: Efficient Triton Kernels for LLM Training},
  author={Hsu, Pin-Lun and Dai, Yun and Kothapalli, Vignesh and Song, Qingquan and Tang, Shao and Zhu, Siyu and Shimizu, Steven and Sahni, Shivam and Ning, Haowen and Chen, Yanning},
  journal={arXiv preprint arXiv:2410.10989},
  year={2024}
}

@article{mohtashami2023landmark,
  title={Landmark attention: Random-access infinite context length for transformers},
  author={Mohtashami, Amirkeivan and Jaggi, Martin},
  journal={arXiv preprint arXiv:2305.16300},
  year={2023}
}

@inproceedings{han2024lm,
  title={LM-Infinite: Zero-Shot Extreme Length Generalization for Large Language Models},
  author={Han, Chi and Wang, Qifan and Peng, Hao and Xiong, Wenhan and Chen, Yu and Ji, Heng and Wang, Sinong},
  booktitle={Proceedings of the 2024 Conference of the North American Chapter of the Association for Computational Linguistics: Human Language Technologies (Volume 1: Long Papers)},
  pages={3991--4008},
  year={2024}
}

@article{peng2023yarn,
  title={Yarn: Efficient context window extension of large language models},
  author={Peng, Bowen and Quesnelle, Jeffrey and Fan, Honglu and Shippole, Enrico},
  journal={arXiv preprint arXiv:2309.00071},
  year={2023}
}

@misc{peng2023ntk,
  title={Ntk-aware scaled rope allows llama models to have extended (8k+) context size without any fine-tuning and minimal perplexity degradation},
  author={Peng, Bowen and Quesnelle, Jeffrey},
  year={2023}
}

@article{chen2023extending,
  title={Extending context window of large language models via positional interpolation},
  author={Chen, Shouyuan and Wong, Sherman and Chen, Liangjian and Tian, Yuandong},
  journal={arXiv preprint arXiv:2306.15595},
  year={2023}
}

@article{team2024gemini,
  title={Gemini 1.5: Unlocking multimodal understanding across millions of tokens of context},
  author={Team, Gemini and Georgiev, Petko and Lei, Ving Ian and Burnell, Ryan and Bai, Libin and Gulati, Anmol and Tanzer, Garrett and Vincent, Damien and Pan, Zhufeng and Wang, Shibo and others},
  journal={arXiv preprint arXiv:2403.05530},
  year={2024}
}

@article{achiam2023gpt,
  title={Gpt-4 technical report},
  author={Achiam, Josh and Adler, Steven and Agarwal, Sandhini and Ahmad, Lama and Akkaya, Ilge and Aleman, Florencia Leoni and Almeida, Diogo and Altenschmidt, Janko and Altman, Sam and Anadkat, Shyamal and others},
  journal={arXiv preprint arXiv:2303.08774},
  year={2023}
}

@misc{jin2024llm,
      title={LLM Maybe LongLM: Self-Extend LLM Context Window Without Tuning}, 
      author={Hongye Jin and Xiaotian Han and Jingfeng Yang and Zhimeng Jiang and Zirui Liu and Chia-Yuan Chang and Huiyuan Chen and Xia Hu},
      year={2024},
      eprint={2401.01325},
      archivePrefix={arXiv},
      primaryClass={cs.CL}
}

@article{ulralong2025,
  title={From 128K to 4M: Efficient Training of Ultra-Long Context Large Language Models},
  author={Xu, Chejian and Ping, Wei and Xu, Peng and Liu, Zihan and Wang, Boxin and Shoeybi, Mohammad and Catanzaro, Bryan},
  journal={arXiv preprint},
  year={2025}
 }

@article{lu2025mobamixtureblockattention,
  author = {Enzhe Lu and Zhejun Jiang and Jingyuan Liu and Yulun Du and Tao Jiang and Chao Hong and Shaowei Liu and Weiran He and Enming Yuan and Yuzhi Wang and Zhiqi Huang and Huan Yuan and Suting Xu and Xinran Xu and Guokun Lai and Yanru Chen and Huabin Zheng and Junjie Yan and Jianlin Su and Yuxin Wu and Yutao Zhang and Zhilin Yang and Xinyu Zhou and Mingxing Zhang and Jiezhong Qiu},
  title = {MoBA: Mixture of Block Attention for Long-Context LLMs},
  journal={arXiv preprint arXiv:2502.13189},
  year={2025}
}

@misc{yuan2025nativesparseattentionhardwarealigned,
      title={Native Sparse Attention: Hardware-Aligned and Natively Trainable Sparse Attention}, 
      author={Jingyang Yuan and Huazuo Gao and Damai Dai and Junyu Luo and Liang Zhao and Zhengyan Zhang and Zhenda Xie and Y. X. Wei and Lean Wang and Zhiping Xiao and Yuqing Wang and Chong Ruan and Ming Zhang and Wenfeng Liang and Wangding Zeng},
      year={2025},
      eprint={2502.11089},
      archivePrefix={arXiv},
      primaryClass={cs.CL},
      url={https://arxiv.org/abs/2502.11089}, 
}

@misc{yang2025qwen251mtechnicalreport,
      title={Qwen2.5-1M Technical Report}, 
      author={An Yang and Bowen Yu and Chengyuan Li and Dayiheng Liu and Fei Huang and Haoyan Huang and Jiandong Jiang and Jianhong Tu and Jianwei Zhang and Jingren Zhou and Junyang Lin and Kai Dang and Kexin Yang and Le Yu and Mei Li and Minmin Sun and Qin Zhu and Rui Men and Tao He and Weijia Xu and Wenbiao Yin and Wenyuan Yu and Xiafei Qiu and Xingzhang Ren and Xinlong Yang and Yong Li and Zhiying Xu and Zipeng Zhang},
      year={2025},
      eprint={2501.15383},
      archivePrefix={arXiv},
      primaryClass={cs.CL},
      url={https://arxiv.org/abs/2501.15383}, 
}

@misc{abdin2024phi4technicalreport,
      title={Phi-4 Technical Report}, 
      author={Marah Abdin and Jyoti Aneja and Harkirat Behl and Sébastien Bubeck and Ronen Eldan and Suriya Gunasekar and Michael Harrison and Russell J. Hewett and Mojan Javaheripi and Piero Kauffmann and James R. Lee and Yin Tat Lee and Yuanzhi Li and Weishung Liu and Caio C. T. Mendes and Anh Nguyen and Eric Price and Gustavo de Rosa and Olli Saarikivi and Adil Salim and Shital Shah and Xin Wang and Rachel Ward and Yue Wu and Dingli Yu and Cyril Zhang and Yi Zhang},
      year={2024},
      eprint={2412.08905},
      archivePrefix={arXiv},
      primaryClass={cs.CL},
      url={https://arxiv.org/abs/2412.08905}, 
}

@misc{yang2025qwen3technicalreport,
      title={Qwen3 Technical Report}, 
      author={An Yang and Anfeng Li and Baosong Yang and Beichen Zhang and Binyuan Hui and Bo Zheng and Bowen Yu and Chang Gao and Chengen Huang and Chenxu Lv and Chujie Zheng and Dayiheng Liu ...},
      year={2025},
      eprint={2505.09388},
      archivePrefix={arXiv},
      primaryClass={cs.CL},
      url={https://arxiv.org/abs/2505.09388}, 
}

@misc{grattafiori2024llama3herdmodels,
      title={The Llama 3 Herd of Models}, 
      author={Aaron Grattafiori and Abhimanyu Dubey and Abhinav Jauhri and Abhinav Pandey and Abhishek Kadian and Ahmad Al-Dahle and Aiesha Letman and Akhil Mathur and Alan Schelten and Alex Vaughan and Amy Yang and Angela Fan...},
      year={2024},
      eprint={2407.21783},
      archivePrefix={arXiv},
      primaryClass={cs.AI},
      url={https://arxiv.org/abs/2407.21783}, 
}

@inproceedings{xiao2024infllm,
  title={Infllm: Training-free long-context extrapolation for llms with an efficient context memory},
  author={Xiao, Chaojun and Zhang, Pengle and Han, Xu and Xiao, Guangxuan and Lin, Yankai and Zhang, Zhengyan and Liu, Zhiyuan and Sun, Maosong},
  booktitle={The Thirty-eighth Annual Conference on Neural Information Processing Systems},
  year={2024}
}

@misc{ni2025teachinglargelanguagemodels,
      title={Teaching Large Language Models to Reason through Learning and Forgetting}, 
      author={Tianwei Ni and Allen Nie and Sapana Chaudhary and Yao Liu and Huzefa Rangwala and Rasool Fakoor},
      year={2025},
      eprint={2504.11364},
      archivePrefix={arXiv},
      primaryClass={cs.LG},
      url={https://arxiv.org/abs/2504.11364}, 
}

@misc{yao2025optimizingchainofthoughtreasonersgradient,
      title={Optimizing Chain-of-Thought Reasoners via Gradient Variance Minimization in Rejection Sampling and RL}, 
      author={Jiarui Yao and Yifan Hao and Hanning Zhang and Hanze Dong and Wei Xiong and Nan Jiang and Tong Zhang},
      year={2025},
      eprint={2505.02391},
      archivePrefix={arXiv},
      primaryClass={cs.LG},
      url={https://arxiv.org/abs/2505.02391}, 
}

@misc{abdin2025phi4reasoningtechnicalreport,
      title={Phi-4-reasoning Technical Report}, 
      author={Marah Abdin and Sahaj Agarwal and Ahmed Awadallah and Vidhisha Balachandran and Harkirat Behl and Lingjiao Chen and Gustavo de Rosa and Suriya Gunasekar and Mojan Javaheripi and Neel Joshi and Piero Kauffmann and Yash Lara and Caio César Teodoro Mendes and Arindam Mitra and Besmira Nushi and Dimitris Papailiopoulos and Olli Saarikivi and Shital Shah and Vaishnavi Shrivastava and Vibhav Vineet and Yue Wu and Safoora Yousefi and Guoqing Zheng},
      year={2025},
      eprint={2504.21318},
      archivePrefix={arXiv},
      primaryClass={cs.AI},
      url={https://arxiv.org/abs/2504.21318}, 
}

@misc{yu2025dapoopensourcellmreinforcement,
      title={DAPO: An Open-Source LLM Reinforcement Learning System at Scale}, 
      author={Qiying Yu and Zheng Zhang and Ruofei Zhu and Yufeng Yuan and Xiaochen Zuo and Yu Yue and Tiantian Fan and Gaohong Liu and Lingjun Liu and Xin Liu and Haibin Lin and Zhiqi Lin and Bole Ma and Guangming Sheng and Yuxuan Tong and Chi Zhang and Mofan Zhang and Wang Zhang and Hang Zhu and Jinhua Zhu and Jiaze Chen and Jiangjie Chen and Chengyi Wang and Hongli Yu and Weinan Dai and Yuxuan Song and Xiangpeng Wei and Hao Zhou and Jingjing Liu and Wei-Ying Ma and Ya-Qin Zhang and Lin Yan and Mu Qiao and Yonghui Wu and Mingxuan Wang},
      year={2025},
      eprint={2503.14476},
      archivePrefix={arXiv},
      primaryClass={cs.LG},
      url={https://arxiv.org/abs/2503.14476}, 
}

@misc{qwq32b,
    title = {QwQ-32B: Embracing the Power of Reinforcement Learning},
    url = {https://qwenlm.github.io/blog/qwq-32b/},
    author = {Qwen Team},
    month = {March},
    year = {2025}
}

@article{qwen2.5,
      title={Qwen2.5 Technical Report}, 
      author={An Yang and Baosong Yang and Beichen Zhang and Binyuan Hui and Bo Zheng and Bowen Yu and Chengyuan Li and Dayiheng Liu and Fei Huang and Haoran Wei and Huan Lin and Jian Yang and Jianhong Tu and Jianwei Zhang and Jianxin Yang and Jiaxi Yang and Jingren Zhou and Junyang Lin and Kai Dang and Keming Lu and Keqin Bao and Kexin Yang and Le Yu and Mei Li and Mingfeng Xue and Pei Zhang and Qin Zhu and Rui Men and Runji Lin and Tianhao Li and Tianyi Tang and Tingyu Xia and Xingzhang Ren and Xuancheng Ren and Yang Fan and Yang Su and Yichang Zhang and Yu Wan and Yuqiong Liu and Zeyu Cui and Zhenru Zhang and Zihan Qiu},
      journal={arXiv preprint arXiv:2412.15115},
      year={2024}
}

@misc{schulman2017proximalpolicyoptimizationalgorithms,
      title={Proximal Policy Optimization Algorithms}, 
      author={John Schulman and Filip Wolski and Prafulla Dhariwal and Alec Radford and Oleg Klimov},
      year={2017},
      eprint={1707.06347},
      archivePrefix={arXiv},
      primaryClass={cs.LG},
      url={https://arxiv.org/abs/1707.06347}, 
}

@article{zhang2024llm,
  title={Llm as a mastermind: A survey of strategic reasoning with large language models},
  author={Zhang, Yadong and Mao, Shaoguang and Ge, Tao and Wang, Xun and de Wynter, Adrian and Xia, Yan and Wu, Wenshan and Song, Ting and Lan, Man and Wei, Furu},
  journal={arXiv preprint arXiv:2404.01230},
  year={2024}
}

@article{zhong2024dpo,
  title={Dpo meets ppo: Reinforced token optimization for rlhf},
  author={Zhong, Han and Shan, Zikang and Feng, Guhao and Xiong, Wei and Cheng, Xinle and Zhao, Li and He, Di and Bian, Jiang and Wang, Liwei},
  journal={arXiv preprint arXiv:2404.18922},
  year={2024}
}

@misc{shao2024deepseekmathpushinglimitsmathematical,
      title={DeepSeekMath: Pushing the Limits of Mathematical Reasoning in Open Language Models}, 
      author={Zhihong Shao and Peiyi Wang and Qihao Zhu and Runxin Xu and Junxiao Song and Xiao Bi and Haowei Zhang and Mingchuan Zhang and Y. K. Li and Y. Wu and Daya Guo},
      year={2024},
      eprint={2402.03300},
      archivePrefix={arXiv},
      primaryClass={cs.CL},
      url={https://arxiv.org/abs/2402.03300}, 
}

@misc{deepseekai2025deepseekr1incentivizingreasoningcapability,
      title={DeepSeek-R1: Incentivizing Reasoning Capability in LLMs via Reinforcement Learning}, 
      author={DeepSeek-AI and Daya Guo and Dejian Yang and Haowei Zhang and Junxiao Song and Ruoyu Zhang...},
      year={2025},
      eprint={2501.12948},
      archivePrefix={arXiv},
      primaryClass={cs.CL},
      url={https://arxiv.org/abs/2501.12948}, 
}

@misc{openr1,
    title = {Open R1: A fully open reproduction of DeepSeek-R1},
    url = {https://github.com/huggingface/open-r1},
    author = {Hugging Face},
    month = {January},
    year = {2025}
}

@article{moshkov2025aimo2,
  title   = {AIMO-2 Winning Solution: Building State-of-the-Art Mathematical Reasoning Models with OpenMathReasoning dataset},
  author  = {Ivan Moshkov and Darragh Hanley and Ivan Sorokin and Shubham Toshniwal and Christof Henkel and Benedikt Schifferer and Wei Du and Igor Gitman},
  year    = {2025},
  journal = {arXiv preprint arXiv:2504.16891}
}

@misc{openthoughts,
  author = {Team, OpenThoughts},
  month = jan,
  title = {{Open Thoughts}},
  howpublished = {https://open-thoughts.ai},
  year = {2025}
}

@misc{deepmath,
      title={DeepMath-103K: A Large-Scale, Challenging, Decontaminated, and Verifiable Mathematical Dataset for Advancing Reasoning}, 
      author={Zhiwei He and Tian Liang and Jiahao Xu and Qiuzhi Liu and Xingyu Chen and Yue Wang and Linfeng Song and Dian Yu and Zhenwen Liang and Wenxuan Wang and Zhuosheng Zhang and Rui Wang and Zhaopeng Tu and Haitao Mi and Dong Yu},
      year={2025},
      eprint={2504.11456},
      archivePrefix={arXiv},
      primaryClass={cs.CL},
      url={https://arxiv.org/abs/2504.11456}, 
}

@article{plaat2024reasoning,
  title={Reasoning with large language models, a survey},
  author={Plaat, Aske and Wong, Annie and Verberne, Suzan and Broekens, Joost and van Stein, Niki and Back, Thomas},
  journal={arXiv preprint arXiv:2407.11511},
  year={2024}
}

@article{team2025kimi,
  title={Kimi k1. 5: Scaling reinforcement learning with llms},
  author={Team, Kimi and Du, Angang and Gao, Bofei and Xing, Bowei and Jiang, Changjiu and Chen, Cheng and Li, Cheng and Xiao, Chenjun and Du, Chenzhuang and Liao, Chonghua and others},
  journal={arXiv preprint arXiv:2501.12599},
  year={2025}
}

@article{guo2025deepseek,
  title={Deepseek-r1: Incentivizing reasoning capability in llms via reinforcement learning},
  author={Guo, Daya and Yang, Dejian and Zhang, Haowei and Song, Junxiao and Zhang, Ruoyu and Xu, Runxin and Zhu, Qihao and Ma, Shirong and Wang, Peiyi and Bi, Xiao and others},
  journal={arXiv preprint arXiv:2501.12948},
  year={2025}
}

@misc{zeng2025simplerlzooinvestigatingtamingzero,
      title={SimpleRL-Zoo: Investigating and Taming Zero Reinforcement Learning for Open Base Models in the Wild}, 
      author={Weihao Zeng and Yuzhen Huang and Qian Liu and Wei Liu and Keqing He and Zejun Ma and Junxian He},
      year={2025},
      eprint={2503.18892},
      archivePrefix={arXiv},
      primaryClass={cs.LG},
      url={https://arxiv.org/abs/2503.18892}, 
}

@misc{sky_t1_2025,
  author       = {NovaSky Team},
  title        = {Sky-T1: Train your own O1 preview model within \$450},
  howpublished = {https://novasky-ai.github.io/posts/sky-t1},
  note         = {Accessed: 2025-01-09},
  year         = {2025}
}

@misc{bespoke_stratos,  
    author = {Bespoke Labs},  
    title = {Bespoke-Stratos: The unreasonable effectiveness of reasoning distillation},  
    howpublished = {www.bespokelabs.ai/blog/bespoke-stratos-the-unreasonable-effectiveness-of-reasoning-distillation},  
    note = {Accessed: 2025-01-22},  
    year = {2025}
}

@misc{li2025llmseasilylearnreason,
      title={LLMs Can Easily Learn to Reason from Demonstrations Structure, not content, is what matters!}, 
      author={Dacheng Li and Shiyi Cao and Tyler Griggs and Shu Liu and Xiangxi Mo and Eric Tang and Sumanth Hegde and Kourosh Hakhamaneshi and Shishir G. Patil and Matei Zaharia and Joseph E. Gonzalez and Ion Stoica},
      year={2025},
      eprint={2502.07374},
      archivePrefix={arXiv},
      primaryClass={cs.AI},
      url={https://arxiv.org/abs/2502.07374}, 
}

@article{ji2025first,
  title={The first few tokens are all you need: An efficient and effective unsupervised prefix fine-tuning method for reasoning models},
  author={Ji, Ke and Xu, Jiahao and Liang, Tian and Liu, Qiuzhi and He, Zhiwei and Chen, Xingyu and Liu, Xiaoyuan and Wang, Zhijie and Chen, Junying and Wang, Benyou and others},
  journal={arXiv preprint arXiv:2503.02875},
  year={2025}
}

@misc{muennighoff2025s1simpletesttimescaling,
      title={s1: Simple test-time scaling}, 
      author={Niklas Muennighoff and Zitong Yang and Weijia Shi and Xiang Lisa Li and Li Fei-Fei and Hannaneh Hajishirzi and Luke Zettlemoyer and Percy Liang and Emmanuel Candès and Tatsunori Hashimoto},
      year={2025},
      eprint={2501.19393},
      archivePrefix={arXiv},
      primaryClass={cs.CL},
      url={https://arxiv.org/abs/2501.19393}, 
}

@misc{ye2025limoreasoning,
      title={LIMO: Less is More for Reasoning}, 
      author={Yixin Ye and Zhen Huang and Yang Xiao and Ethan Chern and Shijie Xia and Pengfei Liu},
      year={2025},
      eprint={2502.03387},
      archivePrefix={arXiv},
      primaryClass={cs.CL},
      url={https://arxiv.org/abs/2502.03387}, 
}

@misc{sui2025stopoverthinkingsurveyefficient,
      title={Stop Overthinking: A Survey on Efficient Reasoning for Large Language Models}, 
      author={Yang Sui and Yu-Neng Chuang and Guanchu Wang and Jiamu Zhang and Tianyi Zhang and Jiayi Yuan and Hongyi Liu and Andrew Wen and Shaochen Zhong and Hanjie Chen and Xia Hu},
      year={2025},
      eprint={2503.16419},
      archivePrefix={arXiv},
      primaryClass={cs.CL},
      url={https://arxiv.org/abs/2503.16419}, 
}

\newpage
\section*{NeurIPS Paper Checklist}

%%% BEGIN INSTRUCTIONS %%%
The checklist is designed to encourage best practices for responsible machine learning research, addressing issues of reproducibility, transparency, research ethics, and societal impact. Do not remove the checklist: {\bf The papers not including the checklist will be desk rejected.} The checklist should follow the references and follow the (optional) supplemental material.  The checklist does NOT count towards the page
limit. 

Please read the checklist guidelines carefully for information on how to answer these questions. For each question in the checklist:
\begin{itemize}
    \item You should answer \answerYes{}, \answerNo{}, or \answerNA{}.
    \item \answerNA{} means either that the question is Not Applicable for that particular paper or the relevant information is Not Available.
    \item Please provide a short (1–2 sentence) justification right after your answer (even for NA). 
   % \item {\bf The papers not including the checklist will be desk rejected.}
\end{itemize}

{\bf The checklist answers are an integral part of your paper submission.} They are visible to the reviewers, area chairs, senior area chairs, and ethics reviewers. You will be asked to also include it (after eventual revisions) with the final version of your paper, and its final version will be published with the paper.

The reviewers of your paper will be asked to use the checklist as one of the factors in their evaluation. While "\answerYes{}" is generally preferable to "\answerNo{}", it is perfectly acceptable to answer "\answerNo{}" provided a proper justification is given (e.g., "error bars are not reported because it would be too computationally expensive" or "we were unable to find the license for the dataset we used"). In general, answering "\answerNo{}" or "\answerNA{}" is not grounds for rejection. While the questions are phrased in a binary way, we acknowledge that the true answer is often more nuanced, so please just use your best judgment and write a justification to elaborate. All supporting evidence can appear either in the main paper or the supplemental material, provided in appendix. If you answer \answerYes{} to a question, in the justification please point to the section(s) where related material for the question can be found.

IMPORTANT, please:
\begin{itemize}
    \item {\bf Delete this instruction block, but keep the section heading ``NeurIPS Paper Checklist"},
    \item  {\bf Keep the checklist subsection headings, questions/answers and guidelines below.}
    \item {\bf Do not modify the questions and only use the provided macros for your answers}.
\end{itemize}

%%% END INSTRUCTIONS %%%

\begin{enumerate}

\item {\bf Claims}
    \item[] Question: Do the main claims made in the abstract and introduction accurately reflect the paper's contributions and scope?
    \item[] Answer: \answerYes{} % Replace by \answerYes{}, \answerNo{}, or \answerNA{}.
    \item[] Justification:  yes, we include the paper’s contributions and scope in the abstract and introduction

\item {\bf Limitations}
    \item[] Question: Does the paper discuss the limitations of the work performed by the authors?
    \item[] Answer: \answerYes{} % Replace by \answerYes{}, \answerNo{}, or \answerNA{}.
    \item[] Justification: we add the limitations to the conclusion

\item {\bf Theory assumptions and proofs}
    \item[] Question: For each theoretical result, does the paper provide the full set of assumptions and a complete (and correct) proof?
    \item[] Answer: \answerYes{} % Replace by \answerYes{}, \answerNo{}, or \answerNA{}.
    \item[] Justification: we include the full set of assumptions and a complete (and correct) proof

    \item {\bf Experimental result reproducibility}
    \item[] Question: Does the paper fully disclose all the information needed to reproduce the main experimental results of the paper to the extent that it affects the main claims and/or conclusions of the paper (regardless of whether the code and data are provided or not)?
    \item[] Answer: \answerYes{} % Replace by \answerYes{}, \answerNo{}, or \answerNA{}.
    \item[] Justification: we include a section about the experimental setup.

\item {\bf Open access to data and code}
    \item[] Question: Does the paper provide open access to the data and code, with sufficient instructions to faithfully reproduce the main experimental results, as described in supplemental material?
    \item[] Answer: \answerYes{} % Replace by \answerYes{}, \answerNo{}, or \answerNA{}.
    \item[] Justification: we provide open access to the data and code

\item {\bf Experimental setting/details}
    \item[] Question: Does the paper specify all the training and test details (e.g., data splits, hyperparameters, how they were chosen, type of optimizer, etc.) necessary to understand the results?
    \item[] Answer: \answerYes{} % Replace by \answerYes{}, \answerNo{}, or \answerNA{}.
    \item[] Justification: we include a section about the experimental setup

\item {\bf Experiment statistical significance}
    \item[] Question: Does the paper report error bars suitably and correctly defined or other appropriate information about the statistical significance of the experiments?
    \item[] Answer: \answerYes{} % Replace by \answerYes{}, \answerNo{}, or \answerNA{}.
    \item[] Justification: we include a stable metric to get results.

\item {\bf Experiments compute resources}
    \item[] Question: For each experiment, does the paper provide sufficient information on the computer resources (type of compute workers, memory, time of execution) needed to reproduce the experiments?
    \item[] Answer: \answerYes{} % Replace by \answerYes{}, \answerNo{}, or \answerNA{}.
    \item[] Justification:: we provide sufficient information on the computer resources.
    
\item {\bf Code of ethics}
    \item[] Question: Does the research conducted in the paper conform, in every respect, with the NeurIPS Code of Ethics \url{https://neurips.cc/public/EthicsGuidelines}?
    \item[] Answer: \answerYes{} % Replace by \answerYes{}, \answerNo{}, or \answerNA{}.
    \item[] Justification: the research conducted in the paper conform with the NeurIPS Code of Ethics

\item {\bf Broader impacts}
    \item[] Question: Does the paper discuss both potential positive societal impacts and negative societal impacts of the work performed?
    \item[] Answer: \answerNA{} % Replace by \answerYes{}, \answerNo{}, or \answerNA{}.
    \item[] Justification: there is no societal impact of the work performed.
    
\item {\bf Safeguards}
    \item[] Question: Does the paper describe safeguards that have been put in place for responsible release of data or models that have a high risk for misuse (e.g., pretrained language models, image generators, or scraped datasets)?
    \item[] Answer: \answerNA{} % Replace by \answerYes{}, \answerNo{}, or \answerNA{}.
    \item[] Justification:  the paper poses no such risks.

\item {\bf Licenses for existing assets}
    \item[] Question: Are the creators or original owners of assets (e.g., code, data, models), used in the paper, properly credited and are the license and terms of use explicitly mentioned and properly respected?
    \item[] Answer: \answerNA{} % Replace by \answerYes{}, \answerNo{}, or \answerNA{}.
    \item[] Justification: the paper does not use existing assets.

\item {\bf New assets}
    \item[] Question: Are new assets introduced in the paper well documented and is the documentation provided alongside the assets?
    \item[] Answer: \answerNA{} % Replace by \answerYes{}, \answerNo{}, or \answerNA{}.
    \item[] Justification: the paper does not release new assets.

\item {\bf Crowdsourcing and research with human subjects}
    \item[] Question: For crowdsourcing experiments and research with human subjects, does the paper include the full text of instructions given to participants and screenshots, if applicable, as well as details about compensation (if any)? 
    \item[] Answer: \answerNA{} % Replace by \answerYes{}, \answerNo{}, or \answerNA{}.
    \item[] Justification:  the paper does not involve crowdsourcing nor research with human subjects.

\item {\bf Institutional review board (IRB) approvals or equivalent for research with human subjects}
    \item[] Question: Does the paper describe potential risks incurred by study participants, whether such risks were disclosed to the subjects, and whether Institutional Review Board (IRB) approvals (or an equivalent approval/review based on the requirements of your country or institution) were obtained?
    \item[] Answer: \answerNA{} % Replace by \answerYes{}, \answerNo{}, or \answerNA{}.
    \item[] Justification: the paper does not involve crowdsourcing nor research with human subjects.

\item {\bf Declaration of LLM usage}
    \item[] Question: Does the paper describe the usage of LLMs if it is an important, original, or non-standard component of the core methods in this research? Note that if the LLM is used only for writing, editing, or formatting purposes and does not impact the core methodology, scientific rigorousness, or originality of the research, declaration is not required.
    %this research? 
    \item[] Answer: \answerNA{} % Replace by \answerYes{}, \answerNo{}, or \answerNA{}.
    \item[] Justification: we do not involve LLMs to impact the methodology.

\end{enumerate}

\newpage
\appendix

\section{Other Results on Different Long-Context Ability}
\label{app:Other Results on Different Long-Context Ability}

This section provides additional experimental results that complement the main findings by examining how different levels of long-context ability influence model performance across multiple reasoning benchmarks.  
While the main paper focused on reasoning accuracy trends, here we provide a broader view including auxiliary benchmarks, output length analysis, and merged-model experiments.

\paragraph{Effect of RoPE scaling.}  
We first analyze the impact of varying \texttt{RoPE $\theta$} scaling factors on \texttt{LLaMA3-8B-Instruct}.  
As shown in \Cref{tab:Effect of RoPE Scaling on Long-Context and Reasoning Performance}, models with stronger long-context ability generally achieve higher reasoning accuracy across \texttt{MATH500}, \texttt{AIME22-24}, and \texttt{GSM8K}.  
The trend is most consistent up to RoPE $\times 16$, after which performance begins to plateau or slightly decline, suggesting diminishing returns when scaling beyond the effective context length of the training data.  
We also observe that output length tends to shrink as accuracy improves, consistent with our analysis in \Cref{sec:appendix_outputlength}, where correct solutions are typically shorter and more concise.

\begin{table}[ht]
\caption{
\textbf{Effect of RoPE $\theta$ Scaling on Long-Context and Reasoning Performance.} 
Evaluation of \texttt{LLaMA3-8B-Instruct} with different \texttt{RoPE $\theta$} scaling factors on the 32K \textit{Needle-in-a-Haystack} benchmark. 
\texttt{Base} refers to the model's performance before SFT, while \texttt{Short} and \texttt{Long} denote results after fine-tuning on short and long reasoning datasets, respectively. \texttt{Avg} represents the average of the \texttt{Short} and \texttt{Long} performances. Length represents the average output length.
}
\label{tab:Effect of RoPE Scaling on Long-Context and Reasoning Performance}
\small{
\resizebox{\textwidth}{!}{
\begin{tabular}{@{}cccccccccc@{}}
\toprule
RoPE                            & 32k long                            & \multicolumn{4}{c}{Acc(\%)}                                                            & \multicolumn{4}{c}{Length}                              \\ \midrule
\multicolumn{1}{c|}{theta}      & \multicolumn{1}{c|}{ctx ability}    & Base           & Short          & Long           & \multicolumn{1}{c|}{Avg}            & Base & Short          & Long           & Avg            \\ \midrule
\multicolumn{10}{c}{MATH500}                                                                                                                                                                                             \\ \midrule
\multicolumn{1}{c|}{$\times1$}  & \multicolumn{1}{c|}{0}              & \textbf{24.40} & 50.68          & 58.92          & \multicolumn{1}{c|}{54.80}          & 746  & 12991          & 11187          & 12089          \\
\multicolumn{1}{c|}{$\times4$}  & \multicolumn{1}{c|}{3.75}           & 20.40          & 52.80          & 61.16          & \multicolumn{1}{c|}{56.98}          & 754  & 9047           & 9949           & 9498           \\
\multicolumn{1}{c|}{$\times8$}  & \multicolumn{1}{c|}{58.30}          & 18.96          & 53.12          & 62.24          & \multicolumn{1}{c|}{57.68}          & 767  & \textbf{8476}  & 9673           & 9075           \\
\multicolumn{1}{c|}{$\times16$} & \multicolumn{1}{c|}{\textbf{77.05}} & 17.28          & \textbf{54.40} & \textbf{64.32} & \multicolumn{1}{c|}{\textbf{59.36}} & 608  & 8782           & 9335           & \textbf{9059}  \\
\multicolumn{1}{c|}{$\times32$}        & \multicolumn{1}{c|}{58.86}          & 15.24          & 53.84          & 61.96          & \multicolumn{1}{c|}{57.90}          & 490  & 8489           & \textbf{9526}  & 9007           \\
\multicolumn{1}{c|}{$\times64$}        & \multicolumn{1}{c|}{35.00}          & 14.20          & 53.28          & 62.36          & \multicolumn{1}{c|}{57.82}          & 453  & 8605           & 9579           & 9092           \\ \midrule
\multicolumn{10}{c}{AIME22-24}                                                                                                                                                                                           \\ \midrule
\multicolumn{1}{c|}{$\times1$}         & \multicolumn{1}{c|}{0}              & 0.22           & 2.67           & 3.11           & \multicolumn{1}{c|}{2.89}           & 1895 & 27482          & 26706          & 27094          \\
\multicolumn{1}{c|}{$\times4$}         & \multicolumn{1}{c|}{3.75}           & \textbf{0.67}  & 3.78           & 5.11           & \multicolumn{1}{c|}{4.45}           & 2048 & 19961          & 20684          & 20322          \\
\multicolumn{1}{c|}{$\times8$}         & \multicolumn{1}{c|}{58.30}          & 0.22           & 2.89           & 6.89           & \multicolumn{1}{c|}{4.89}           & 1723 & \textbf{17532} & \textbf{17547} & \textbf{17539} \\
\multicolumn{1}{c|}{$\times16$}        & \multicolumn{1}{c|}{\textbf{77.05}} & 0.22           & \textbf{4.67}  & 6.22           & \multicolumn{1}{c|}{\textbf{5.45}}  & 1617 & 18339          & 18015          & 18177          \\
\multicolumn{1}{c|}{$\times32$}        & \multicolumn{1}{c|}{58.86}          & 0.00           & 3.56           & \textbf{7.11}  & \multicolumn{1}{c|}{5.34}           & 1372 & 17876          & 18071          & 17973          \\
\multicolumn{1}{c|}{$\times64$}        & \multicolumn{1}{c|}{35.00}          & 0.44           & 3.11           & 4.67           & \multicolumn{1}{c|}{3.89}           & 1108 & 17830          & 18251          & 18041          \\ \midrule
\multicolumn{10}{c}{GSM8K}                                                                                                                                                                                               \\ \midrule
\multicolumn{1}{c|}{$\times1$}         & \multicolumn{1}{c|}{0}              & \textbf{78.13} & 84.81          & 84.64          & \multicolumn{1}{c|}{84.73}          & 195  & 2244           & 4769           & 3506           \\
\multicolumn{1}{c|}{$\times4$}         & \multicolumn{1}{c|}{3.75}           & 75.54          & 85.64          & 85.88          & \multicolumn{1}{c|}{85.76}          & 190  & \textbf{1973}  & 4010           & 2992           \\
\multicolumn{1}{c|}{$\times8$}         & \multicolumn{1}{c|}{58.30}          & 73.03          & \textbf{85.90} & 85.79          & \multicolumn{1}{c|}{85.85}          & 184  & 1981           & \textbf{3812}  & \textbf{2897}  \\
\multicolumn{1}{c|}{$\times16$}        & \multicolumn{1}{c|}{\textbf{77.05}} & 70.39          & 85.64          & \textbf{86.38} & \multicolumn{1}{c|}{\textbf{86.01}} & 182  & 2019           & 3936           & 2977           \\
\multicolumn{1}{c|}{$\times32$}        & \multicolumn{1}{c|}{58.86}          & 66.35          & 85.41          & 86.14          & \multicolumn{1}{c|}{85.78}          & 180  & 2026           & 4179           & 3103           \\
\multicolumn{1}{c|}{$\times64$}   & \multicolumn{1}{c|}{35.00}                               & 62.21          & 85.56          & 85.25          & 85.41                               & 182  & 2083           & 4177           & 3130           \\ \bottomrule
\end{tabular}
}
}
\end{table}

\paragraph{Extremely long context via model merging.}  
To test whether extremely long contexts (e.g., 128k or 1M) provide further gains, we construct merged variants of \texttt{Qwen2.5-7B-Instruct} by combining its base 32k model with the ultra-long version \texttt{Qwen2.5-7B-Instruct-1M}.  
By adjusting merge ratios, we obtain intermediate models with controllable long-context capabilities, while preserving base reasoning ability.  
As shown in \Cref{fig:Visualization of the relationship between 128k long-context ability}, reasoning performance correlates with effective long-context strength: models with moderate merge ratios (e.g., 0.1, 0.7) achieve consistently strong accuracy, whereas the pure 1M model shows weaker effective long-context utilization and degraded reasoning.

\section{Other Results on Different Merging Ratios}
\label{app:other-merging}

We further analyze how different merging ratios between base and ultra-long variants affect both retrieval ability and downstream reasoning. This provides insight into whether extremely long contexts (e.g., 1M tokens) always yield benefits, or whether moderate ratios strike a better balance between long-context integration and base reasoning stability.

\begin{table}[ht]
\caption{
\textbf{Reasoning Performance After SFT with Different Merge Ratios.} 
Qwen results are based on merging \texttt{Qwen2.5-7B-Instruct-1M} with \texttt{Qwen2.5-7B-Instruct}. A ratio of 0.1 means that the long-context (1M) variant contributes 10\% to the merged model. 
}
\label{tab:Reasoning Performance After SFT with Different Merge Ratios}
\resizebox{\textwidth}{!}{
\begin{tabular}{@{}cccccccccc@{}}
\toprule
\multicolumn{1}{c|}{1M Merge} & \multicolumn{1}{c|}{32k long}   & \multicolumn{4}{c|}{Acc(\%)}                                                           & \multicolumn{4}{c}{Length}            \\ \cmidrule(l){3-10} 
\multicolumn{1}{c|}{Ratio}    & \multicolumn{1}{c|}{ctx ability} & Base           & Short          & Long           & \multicolumn{1}{c|}{Avg}            & Base & Short & Long  & Avg            \\ \midrule
\multicolumn{10}{c}{MATH500}                                                                                                                                                                      \\ \midrule
0   & 78.1 & \textbf{75.00} & 83.48 & 84.84 & 84.16 & 638  & 5572  & 6545  & 6058 \\
0.1 & 79.1 & 74.64 & 83.56 & \textbf{86.28} & \textbf{84.92} & 670 & 5625 & 6432 & \textbf{6028} \\
0.7 & 79.5 & 72.80 & \textbf{83.88} & 84.88 & 84.38 & 630 & 5812 & 6563 & 6187 \\
1.0 & 77.7 & 72.16 & 82.28 & 83.48 & 82.88 & 688 & 5897 & 6881 & 6389 \\ \midrule
\multicolumn{10}{c}{AIME22-24} \\ \midrule
0   & 78.1 & 8.22 & 17.78 & 23.33 & 20.56 & 1051 & 19642 & 18592 & 19117 \\
0.1 & 79.1 & \textbf{9.33}  & 19.11 & 22.89 & 21.00 & 1313 & 19473 & 18178 & \textbf{18825} \\
0.7 & 79.5 & 7.78 & \textbf{19.56} & \textbf{23.56} & \textbf{21.56} & 1416 & 20396 & 18142 & 19269 \\
1.0 & 77.7 & 7.33 & 16.22 & 20.00 & 18.11 & 1750 & 20507 & 18680 & 19594 \\ \midrule
\multicolumn{10}{c}{GSM8K} \\ \midrule
0   & 78.1 & \textbf{90.75} & \textbf{93.19} & 91.54 & 92.37 & 259  & 1429  & 3264  & 2347 \\
0.1 & 79.1 & \textbf{90.75} & 93.15 & 92.16 & 92.66 & 252  & 1472  & 3197  & 2334 \\
0.7 & 79.5 & 89.1 & 93.07 & \textbf{92.27} & \textbf{92.67} & 254  & 1488  & 2930  & 2209 \\
1.0 & 77.7 & 89.13 & 92.84 & 92.34 & 92.59 & 254  & 1494  & 2915  & \textbf{2204} \\ \bottomrule
\end{tabular}
}
\end{table}

\begin{figure}[t]
    \centering
    \begin{subfigure}{0.45\textwidth}
        \centering
        \includegraphics[width=\linewidth]{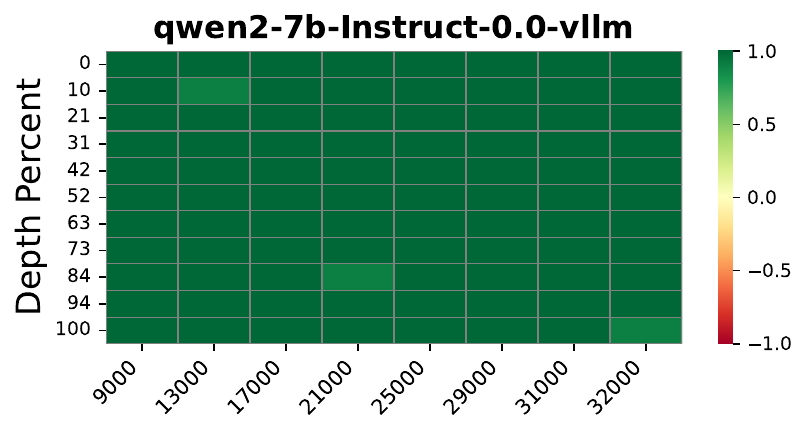}
    \end{subfigure}
    \begin{subfigure}{0.45\textwidth}
        \centering
        \includegraphics[width=\linewidth]{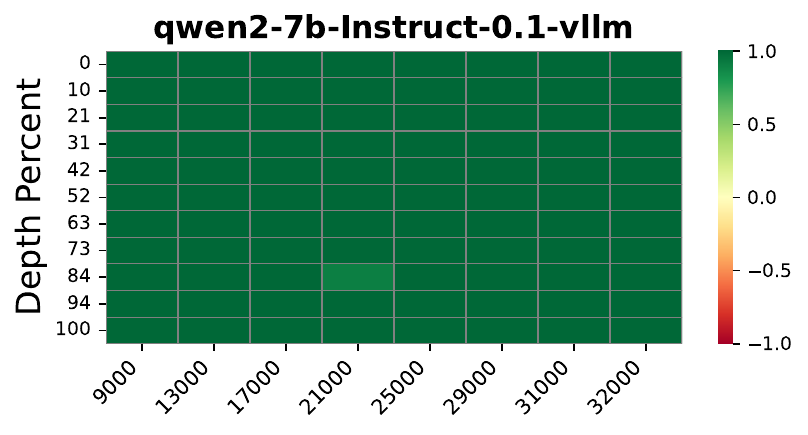}
    \end{subfigure}
    \begin{subfigure}{0.45\textwidth}
        \centering
        \includegraphics[width=\linewidth]{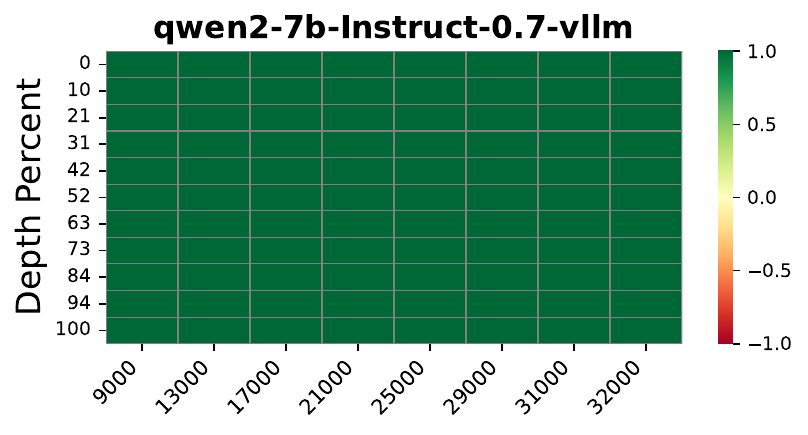}
    \end{subfigure}
    \begin{subfigure}{0.45\textwidth}
        \centering
        \includegraphics[width=\linewidth]{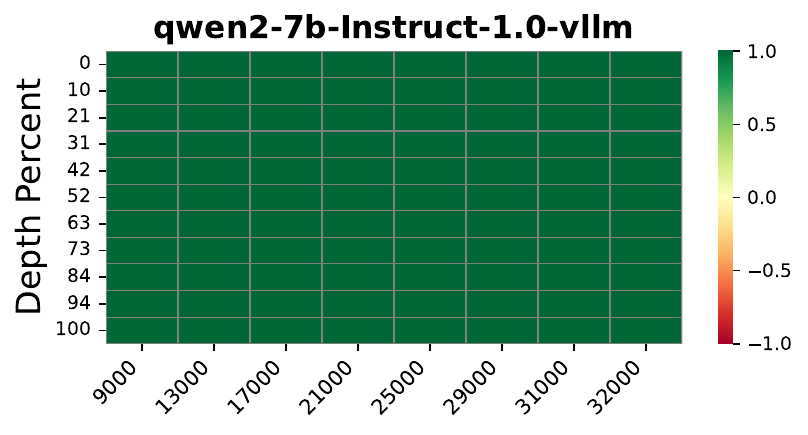}
    \end{subfigure}
    \vspace{-8pt}
    \caption{32K Needle-in-a-Haystack evaluation for different merge ratios. Models with higher long-context contribution achieve stronger retrieval ability, but performance differences saturate quickly.}
    \label{fig:needle-mergeresults}
\end{figure}

\paragraph{Needle-in-a-Haystack (retrieval ability).}  
We first evaluate the merged models on the 32K \textit{Needle-in-a-Haystack} benchmark.  
As shown in \Cref{fig:needle-mergeresults}, long-context ability generally increases with higher contribution from the 1M-token variant.  
However, because most models achieve near-perfect scores, the benchmark does not fully differentiate their effective long-context strength.  
This motivates evaluating more challenging reasoning datasets where differences manifest more clearly.

\paragraph{Reasoning benchmarks.}  
We then fine-tune the merged models on both short and long reasoning datasets and evaluate them on three benchmarks: \texttt{MATH500}, \texttt{AIME22-24}, and \texttt{GSM8K}.  
\Cref{tab:Reasoning Performance After SFT with Different Merge Ratios} reports accuracy and output length.  
Three key observations emerge:  
(1) Moderate merge ratios (e.g., 0.1 or 0.7) yield the best overall reasoning performance, consistently outperforming both the base (0) and fully merged (1.0) variants.  
(2) Pure 1M models (ratio 1.0) exhibit degraded reasoning accuracy, despite higher nominal long-context capacity, suggesting weaker effective utilization.  
(3) Output lengths tend to shrink as reasoning improves, consistent with the observation that successful reasoning requires fewer redundant tokens.

% \paragraph{Output length trends.}  
% We also analyze the relationship between long-context ability and output length (\Cref{fig:length-mergeresults}).  
% For all benchmarks, moderate merge ratios (0.1–0.7) reduce output verbosity while maintaining or improving accuracy, suggesting more concise and efficient reasoning.  
% In contrast, the fully merged 1.0 models often produce unnecessarily long outputs, reflecting weaker alignment between long-context memory and reasoning efficiency.

\begin{table}[t]
\centering
\caption{Performance of \texttt{Phi-4} under different RoPE scales on AIME.}
\label{tab:phi4-aime}
\begin{tabular}{l|c|c|c|c}
\toprule
Model & RoPE & 32k NIAH (\%) & AIME (before, \%) & AIME (after, \%) \\
\midrule
Phi-4 & $\times$1  & 52.27 & 14.89 & 47.56 \\
Phi-4 & $\times$4  & 78.07 & 13.78 & 49.45 \\
Phi-4 & $\times$16 & 84.77 & 10.22 & 50.17 \\
\bottomrule
\end{tabular}
\end{table}

% \begin{figure}[t]
% \centering
%     \begin{subfigure}{0.32\textwidth}
%         \centering
%         \includegraphics[width=\linewidth]{pics/qwen2_math500_len.pdf}
%         \caption{MATH500}
%     \end{subfigure}
%     \begin{subfigure}{0.32\textwidth}
%         \centering
%         \includegraphics[width=\linewidth]{pics/qwen2_aime_len.pdf}
%         \caption{AIME22-24}
%     \end{subfigure}
%     \begin{subfigure}{0.32\textwidth}
%         \centering
%         \includegraphics[width=\linewidth]{pics/qwen2_gsm8k_len.pdf}
%         \caption{GSM8K}
%     \end{subfigure}
%     \vspace{-5pt}
%     \caption{Average output length across different merge ratios of \texttt{Qwen2.5-7B-Instruct}. Moderate ratios yield shorter, more concise reasoning outputs while maintaining strong accuracy.}
%     \label{fig:length-mergeresults}
% \end{figure}

% \paragraph{Takeaways.}  
% These results suggest that extremely long-context capacity (1M) is not always beneficial for reasoning.  
% Instead, moderate integration ratios (0.1–0.7) achieve the best trade-off, balancing long-context retrieval and concise reasoning.  
% This highlights the importance of tuning context capacity to align with reasoning objectives, rather than maximizing nominal window size.

\section{Additional Analysis on Output Length and Context Extension}
\label{sec:appendix-length}

\paragraph{Why is the output length decreasing with the long context extension?}  
The observed decrease in output length is actually aligned with the ratio of correct answers. Correct generations tend to be much shorter and more concise than incorrect ones. With long-context extension, the number of correct answers increases, thus reducing the overall average output length. This phenomenon has also been discussed in prior work \citep{sui2025stopoverthinkingsurveyefficient}.

To support this explanation, we report the average lengths of correct and incorrect generations across both short (0–8k) and long (8–16k) training settings for \texttt{LLaMA3-8B} under different RoPE configurations: \textit{RoPE $\times 1$ (8k context)} and \textit{RoPE $\times 16$ (Extended context)}

\begin{table}[ht]
\centering
\caption{Effect of training length on output length under different RoPE scales.}
\label{tab:rope-length}
\begin{tabular}{l|lcccccc}
\toprule
RoPE Scale & Setting & Accuracy (\%) & Avg Length & \# Correct & Avg & \# Wrong & Avg \\
\midrule
$\times$1  & short & 50.68 & 12991 & 1267 & 3189 & 1233 & 23145 \\
$\times$1  & long  & 58.92 & 11187 & 1473 & 5928 & 1027 & 20999 \\
\midrule
$\times$16 & short & 54.40 & 8782  & 1360 & 3906 & 1140 & 14921 \\
$\times$16 & long  & 64.32 & 9335  & 1608 & 6063 & 892  & 17265 \\
\bottomrule
\end{tabular}
\end{table}

As shown, models with longer context capacity (e.g., RoPE $\times 16$) achieve higher accuracy while generating shorter outputs on average—primarily because a larger portion of the outputs are correct, and correct answers are generally shorter.

\end{document}